\documentclass[conference]{IEEEtran}
\IEEEoverridecommandlockouts
\usepackage{pgfplots}
\pgfplotsset{compat=1.18}
\definecolor{darkgreen}{rgb}{0,0.5,0}
\usepackage{amsmath,amssymb,amsfonts}
\usepackage{algorithmic}
\usepackage{graphicx}
\usepackage{textcomp}
\usepackage{xcolor}

\usepackage[
  backend=biber,
  style=ieee,
  sorting=none
]{biblatex}
\usepgfplotslibrary{fillbetween}
\def\BibTeX{{\rm B\kern-.05em{\sc i\kern-.025em b}\kern-.08em
    T\kern-.1667em\lower.7ex\hbox{E}\kern-.125emX}}
\begin{document}

\title{Spatial Graph Representation and Morphometric Analysis of the Pulmonary Vascular Tree From Computed Tomography Using Multi-Scale Hessian-Based Filter Fusion and TEASAR Skeletonization\\

}

\author{\IEEEauthorblockN{1\textsuperscript{st} Piotr Mackiewicz}
\IEEEauthorblockA{\textit{Faculty of Electrical Engineering} \\
\textit{Warsaw University of Technology}\\
Warsaw, Poland}
\and
\IEEEauthorblockN{2\textsuperscript{nd} Jakub Kołyska}
\IEEEauthorblockA{\textit{Faculty of Electrical Engineering} \\
\textit{Warsaw University of Technology}\\
Warsaw, Poland \\}
\and
\IEEEauthorblockN{3\textsuperscript{rd} Radosław Roszczyk}
\IEEEauthorblockA{\textit{Faculty of Electrical Engineering} \\
\textit{Warsaw University of Technology}\\
Warsaw, Poland \\
radoslaw.roszczyk@pw.edu.pl}

}

\maketitle

\begin{abstract}
Reconstructing the pulmonary vascular tree from computed tomography (CT) images is essential for quantitative lung analysis, vascular morphology assessment, and patient-specific modeling, but remains challenging because pulmonary vessels span multiple spatial scales, from proximal arteries to distal microvasculature. This difficulty is further increased by limited CT scanner resolution, heterogeneous image quality, partial-volume effects, and respiratory motion artifacts in clinical chest CT acquisitions. In contrast to deep learning-based pulmonary vessel segmentation methods that require large annotated datasets, we propose a deterministic, training-free, and fully explainable pipeline for pulmonary vascular tree reconstruction directly from CT data. The method combines multiscale Hessian-based Frangi and Sato vesselness filters using a weighted maximum response across 12 spatial scales ranging from 1 to 8 mm, enabling detection of both large pulmonary arteries and fine peripheral vascular branches. Lung parenchyma segmentation is performed using Hounsfield-unit thresholding, morphological post-processing, and Chan-Vese active-contour refinement. Vascular centerlines are then extracted with the Kimimaro implementation of the TEASAR algorithm, and separate left and right lung vascular graphs are constructed, pruned to remove spurious branches, and validated for acyclicity. The anatomical and geometric plausibility of the reconstructed pulmonary vascular trees is evaluated using volumetric fractal dimension, Strahler order analysis, Horton ratios, and Murray’s law. The resulting volumetric fractal dimension of approximately 2.3 is consistent with values reported for the human pulmonary vasculature, while deviations in Strahler orders and Horton ratios reflect the expected truncation of distal vessels caused by finite CT resolution. These results indicate that the proposed explainable CT-based vascular reconstruction pipeline can generate geometrically plausible pulmonary vascular tree models and may support future applications in quantitative pulmonary imaging, vascular morphology analysis, and computational lung modeling.
\end{abstract}


\section{Introduction}
Accurate reconstruction and quantitative modeling of the pulmonary vascular tree are essential for the diagnosis and longitudinal monitoring of pulmonary diseases in which alterations in vascular morphology have direct clinical relevance~\cite{helmberger2014}. Graph-based descriptors of vascular architecture, including fractal dimension, branching ratios, Strahler orders, and radius-scaling relationships, provide objective morphometric biomarkers that complement visual assessment and may support earlier detection of vascular remodeling.

Robust extraction of these descriptors from clinical CT data, however, remains challenging. The pulmonary vasculature spans a broad range of vessel diameters, from the main pulmonary arteries to submillimeter peripheral branches, while acquisition noise, limited spatial resolution, partial-volume effects, and respiratory motion artifacts reduce vessel conspicuity. Deep-learning segmentation methods have achieved promising results, but typically require large voxel-level annotated datasets that are expensive to produce, sensitive to scanner and protocol variability, and difficult to reproduce across clinical sites. In this work, we therefore present a preliminary deterministic and training-free alternative for CT-based pulmonary vascular tree reconstruction, and assess the resulting vascular graphs using established morphometric and physiological criteria.
\section{Materials}
\label{sec:materials}

\subsection{Computed Tomography Data Acquisition}
\label{ssec:ct_data}
The dataset consists of a single non-contrast chest computed tomography (CT) examination
acquired on a Siemens Symbia~T6 hybrid SPECT/CT scanner (Siemens Healthineers, Erlangen, Germany) running syngo~CT~2007E software (version VA61B), consistent with the hybrid SPECT/CT acquisition setting previously used for lung volume based perfusion CT comparison in pulmonary embolism analysis~\cite{Muchorowski2025}. Acquisition parameters were: tube voltage 130\,kV, tube current-time product 99\,mAs (40\,mA exposure), reconstruction kernel B30s (medium soft-tissue), 214 axial slices at 2.5\,mm slice thickness with in-plane pixel spacing of $0.98 \times 0.98$\,mm on a $512 \times 512$ matrix. As the CT component of this hybrid system is designed for attenuation correction rather than diagnostic imaging, the dataset presents elevated noise and limited resolution compared to a dedicated multidetector CT.

\section{Methods}
\label{sec:methods}

\subsection{Preprocessing: DICOM Load and Isotropic Resampling}
\label{ssec:preprocessing}
Digital Imaging and Communications in Medicine (DICOM) files are loaded with pydicom~\cite{pydicom} and slices are converted to Hounsfield Units (HU). The volume is resampled to an isotropic $1.0 \times 1.0 \times 1.0$\,mm spacing using trilinear interpolation (\texttt{scipy.ndimage.zoom}~\cite{scipy}). Isotropic voxels are essential for the subsequent Hessian-based vesselness filters, which assume equal resolution in all three axes, as well as distance measurements (EDT, branch lengths) throughout the pipeline.

\subsection{Lung Segmentation}
\label{ssec:lung_segmentation}

Lung parenchyma was isolated by HU thresholding ($-1000 \leq \text{HU} \leq -400$) within a body mask, followed by morphological erosion with a spherical structuring element of radius~2 voxels to disconnect the trachea and main bronchi, and retention of the two largest connected components, the left and right lungs. After morphological closure (ball radius~2) and hole filling, the boundary was refined using a morphological Chan--Vese active contour~\cite{chan2001} operating at a reduced resolution of 2.5\,mm ($\text{num\_iter} = 35$, $\text{smoothing} = 2$, $\lambda_1 = \lambda_2 = 1$). Finally, voxels within 3.0\,mm of the lung boundary were excluded via the Euclidean distance transform to suppress pleural-wall artifacts.

\begin{figure}
    \centering
    \includegraphics[width=0.8\linewidth]{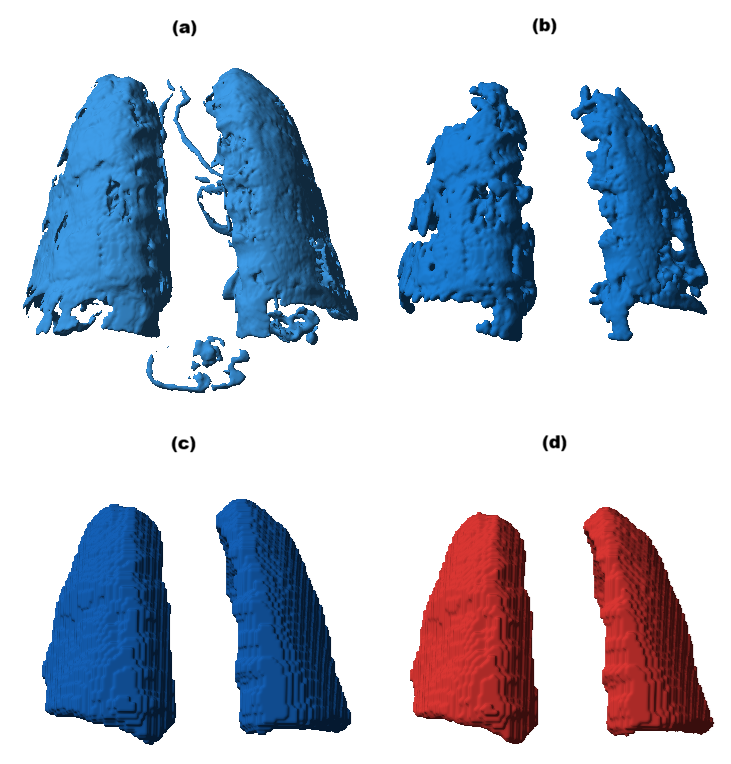}
    \caption{Four stages of lung segmentation pipeline, shown as 3D surfaces of the corresponding binary masks. (a)~HU thresholding, (b)~Airway disconnection, (c)~Chan--Vese refinement, (d)~EDT exclusion.}
      \label{fig:segmentation}
  \end{figure}
    
\subsection{Multi-Scale Vesselness Filter}
\label{ssec:vesselness}
Prior to vesselness computation, the HU volume was smoothed using 3D Perona--Malik anisotropic diffusion~\cite{perona1990} (5~iterations, $\kappa = 50$) to reduce acquisition noise while preserving vessel boundaries. Pulmonary vessels span a wide range of diameters, from large proximal arteries to fine peripheral branches. To capture this range, Hessian-based vesselness filters were applied across 12 spatial scales ($\sigma = 1.0, 1.5, 2.0, 2.5, 3.0, 3.5, 4.0, 4.5, 5.0, 5.5, 7.0, 8.0$~mm), with a sparser sampling above 5.5~mm reflecting the broadening of the Hessian response peak at larger $\sigma$. At each scale, the Hessian matrix of the smoothed HU volume is decomposed into eigenvalues, whose configuration distinguishes tubular structures from planar or blob-like ones. 

Two complementary filters were computed, both implemented in
scikit-image~\cite{vanderwalt2014scikit}. The Frangi filter~\cite{frangi1998} was configured with plate suppression $\alpha = 0.5$, blob suppression $\beta = 0.5$ and structuredness threshold $\gamma = 15$. The Sato filter~\cite{sato1998} was applied with its default eigenvalue weighting ($\alpha_1 = 0.5$, $\alpha_2 = 2.0$). Each filter responds differently to vessel contrast and background noise, providing complementary sensitivity across the vascular tree.

Prior to fusion, each filter map was independently normalized to $[0,\,1]$ by dividing by the 99.9th percentile of non-zero values within the eroded lung mask. The two normalized maps $F_{\mathrm{norm}}$, $S_{\mathrm{norm}}$ were then fused by a weighted voxel-wise maximum,
\begin{equation}
V(x) = \max\!\bigl(w_{\mathrm{F}}\,F_{\mathrm{norm}}(x),\ w_{\mathrm{S}}\,S_{\mathrm{norm}}(x)\bigr),
\label{eq:fusion}
\end{equation}
with $w_{\mathrm{F}} = 1.0$ and $w_{\mathrm{S}} = 1.2$, so that at each voxel the stronger filter response is retained, with a slight preference for the Sato map to favor its response on medium to large vessels. Finally, a noise floor was applied by zeroing all values below the 30th percentile of the non-zero voxels of~$V$, followed by rescaling to $[0,\,1]$ using the 99.5th percentile with clipping. The resulting continuous vesselness map is passed to the skeleton extraction stage.

\begin{figure}
    \centering
    \includegraphics[width=0.9\linewidth]{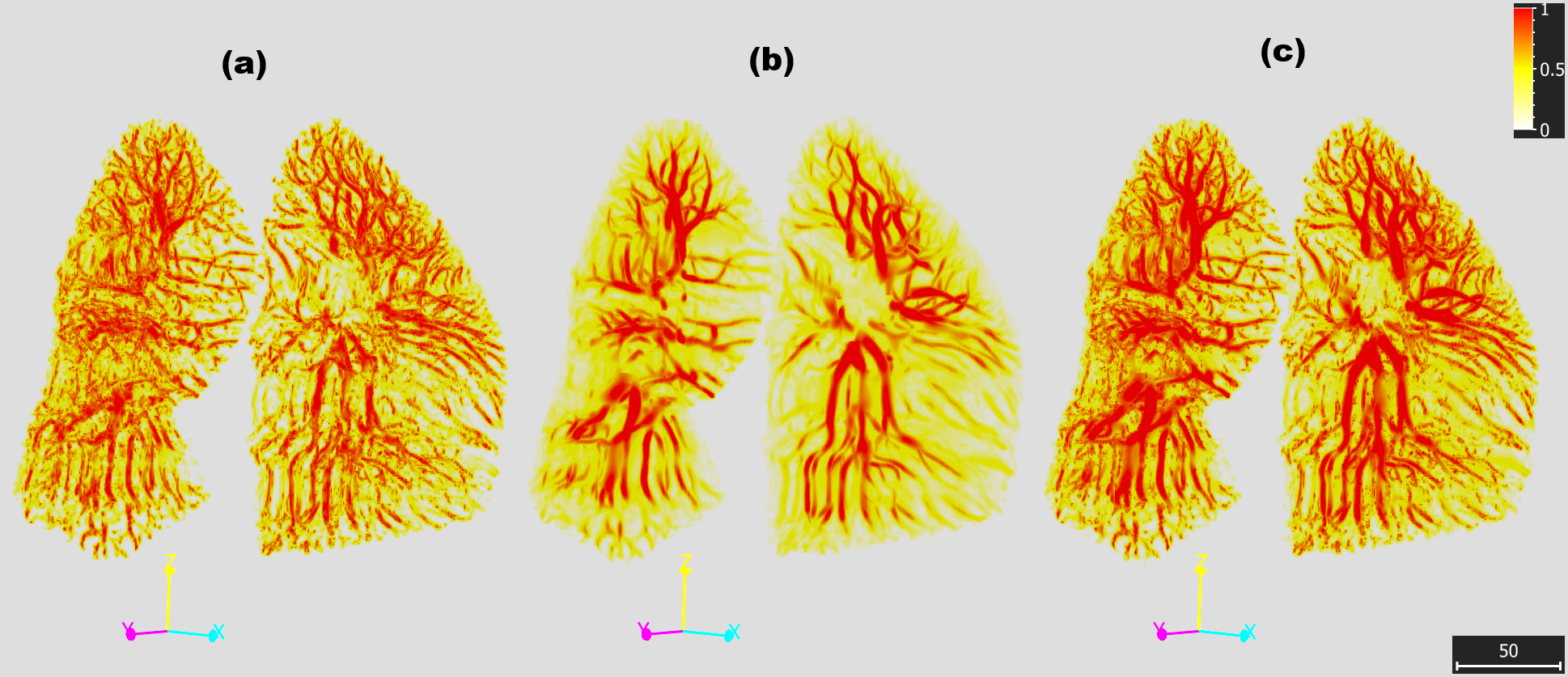}
    \caption{Side-by-side comparison of the two base vesselness responses and their fusion. (a)~\textbf{Frangi}: performs well at highlighting thin, distal branches and preserves a dense network of peripheral vessels close to the pleura. However, it struggles to maintain continuity along the larger arteries, which occasionally appear fragmented rather than continuous tubes. (b)~\textbf{Sato}: performs noticeably worse on the smallest distal branches but reconstructs the larger central arteries much more reliably. (c)~\textbf{Fused}: fusion combines Frangi's sensitivity to small peripheral vessels with Sato's continuity on large central ones, producing a single probability field in which both the main pulmonary arteries and the distal branches are simultaneously better represented.}
    \label{fig:vesselness_compare3}
\end{figure}

  \begin{figure}
      \centering
      \includegraphics[width=0.9\linewidth]{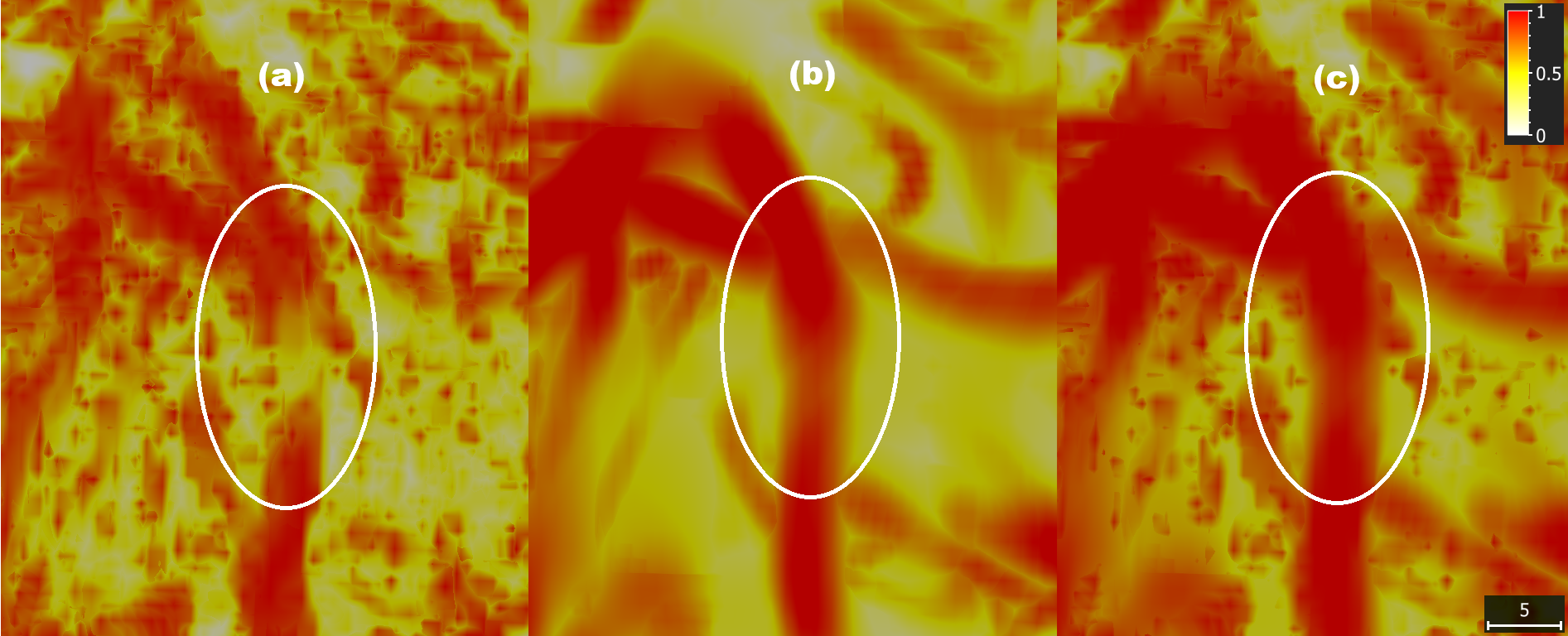}
      \caption{Close-up on a medium pulmonary vessel. (a)~Under \textbf{Frangi}, the vessel is disconnected, reflecting the filter's difficulty with larger, locally non-ideal tubular structures. (b)~Under \textbf{Sato}, the same vessel appears as a continuous tube. (c)~In the \textbf{fused} map, the gaps left by Frangi are filled by the Sato response, restoring a single uninterrupted vessel.}
      \label{fig:placeholder}
  \end{figure}
  
\subsection{Skeleton Extraction}
\label{ssec:skeleton}
The fused vesselness map is binarized at 0.15 and processed independently for each lung. Centerline extraction is performed using TEASAR~\cite{b8}, as implemented in Kimimaro~\cite{b_kimimaro}. It traces geodesic paths on a penalized distance field and invalidates voxels within a data-dependent radius around each accepted path, producing a loop-free centerline with a per-vertex radius estimate. 

For each lung, the Euclidean distance transform (EDT) of the vessel mask serves as both the path-cost field and the source of vessel radii. The path cost uses a penalized distance-to-boundary field (DBF) with $\text{pdrf\_scale}=5000$ and exponent $4$, strongly favoring centerline routes. The invalidation radius is set to $1.5\cdot\text{DBF}+0.5$~mm, with clean bifurcations enforced and voxel anisotropy handled in physical units. Fragments shorter than five voxels are discarded.

\subsection{Graph Construction}
\label{ssec:graph_construction}
The skeleton vertices and edges produced by TEASAR are converted into an undirected graph (NetworkX~\cite{b_networkx}) for each lung independently. Degree-two vertices, which represent sample points along unbranched vessel segments, are compressed so that each graph edge spans an entire branch from one junction to the next (or to a terminal point). Each edge stores the ordered coordinate path along the branch, the integrated arc length $\ell_{\mathrm{mm}}$, and the number of vertices. For duplicate edges with the same endpoints, only the shorter one is kept to avoid parallel graph edges.
 
As pulmonary vascular trees are topologically acyclic, any remaining cycle is attributed to skeletonization artifacts. The cycle count $\beta_1 = E - V + C$ is evaluated per connected component, and for every component with $\beta_1 > 0$, a maximum spanning tree is extracted using $\ell_{\mathrm{mm}}$ as the edge weight, preserving the longest branches.

The resulting forest is cleaned by iteratively pruning degree-one branches shorter than $\ell_{\min} = 2.0$~mm and removing components with fewer than five nodes. Each node is classified as \emph{endpoint} ($\deg \leq 1$), \emph{path} ($\deg = 2$), or \emph{junction} ($\deg \geq 3$). $\beta_1$ is then re-evaluated to confirm that the final graph is a proper forest. The implementation of the proposed reconstruction pipeline, including preprocessing, vesselness filtering, skeleton extraction, and graph analysis scripts, is available at: \url{https://github.com/rroszczyk/PulmoVesselGraph}.

\begin{figure}
    \centering
    \includegraphics[width=0.9\linewidth]{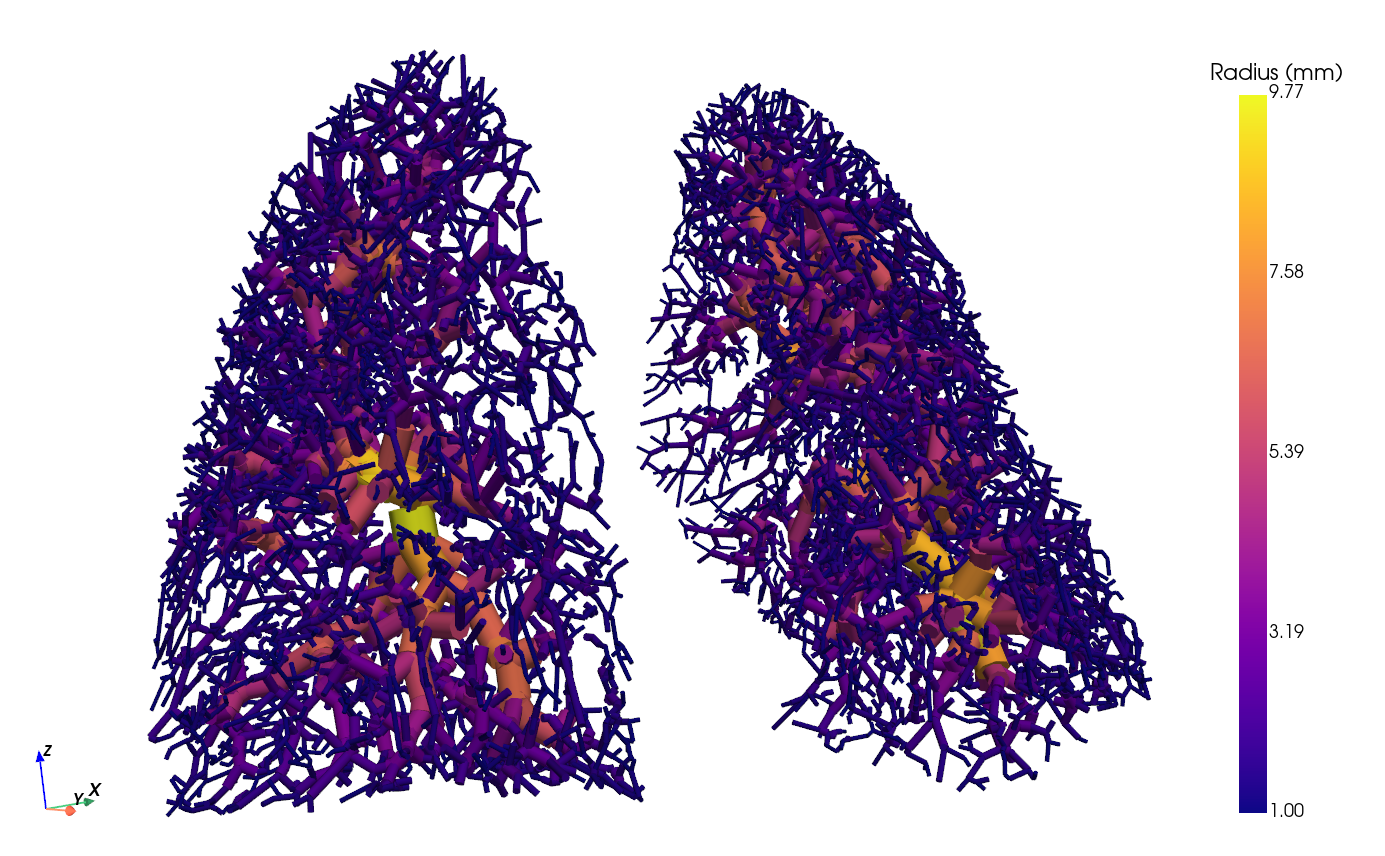}
    \caption{Final pulmonary vascular graph reconstructed. }
    \label{fig:graphpul}
\end{figure}

\subsection{Morphometric Analysis}
\label{ssec:morphometry}

  The reconstructed vascular graph is validated against established
  morphometric laws of branching networks. All metrics are computed
  independently for the left and right lungs.

\subsubsection{Fractal Dimension}

  The space-filling complexity of the vascular tree is quantified via
  3D box-counting~\cite{mandelbrot1982}. The volume is
  partitioned into cubes of side length~$\epsilon$, and the number of
  non-empty boxes $N(\epsilon)$ is recorded for
  $\epsilon \in \{2,4,8,16,32,64\}$~voxels. The fractal dimension is
  obtained from linear regression in log--log space:
  \begin{equation}
    D_f = -\frac{\mathrm{d}\,\ln N(\epsilon)}{\mathrm{d}\,\ln \epsilon}\,.
    \label{eq:fractal}
  \end{equation}
  Two variants are reported: $D_f^{\mathrm{skel}}$ computed on the
  one-voxel skeleton, and $D_f^{\mathrm{vol}}$ on the skeleton inflated
  by the per-node EDT radius, which captures vessel caliber.

\subsubsection{Murray's Law}
\label{subsec:murray}
This law~\cite{murray1926} states that a parent vessel of radius $r_0$ divides into daughter vessels $r_1, r_2, \ldots$ such that
\begin{equation}
    r_0^{\,\alpha} = \sum_{i} r_i^{\,\alpha}\,,
    \label{eq:murray}
\end{equation}
where the classical optimum $\alpha = 3$ minimizes the combined cost of viscous power dissipation and metabolic vessel maintenance under laminar flow~\cite{sherman1981}; recent measurements in human pulmonary arteries report $\alpha = 2.31 \pm 0.60$~\cite{revisiting_murray2025}. The exponent is estimated by bounded nonlinear minimization of
\begin{equation}
    \mathcal{L}(\alpha) = \sum_{j} \Bigl[\,\alpha \log r_{0,j} - \log\!\Bigl(\sum_{i} r_{i,j}^{\,\alpha}\Bigr)\Bigr]^{2}
    \label{eq:murray_objective}
\end{equation}
over all bifurcation nodes of degree~$\geq 3$, with the largest mean radius neighbor taken as the parent.

\subsubsection{Strahler Ordering}
  Each edge is assigned a Strahler order~\cite{strahler1957} via
  post-order traversal from the root node (identified as the node with
  the largest EDT radius). Leaves receive order~1. At a junction, the
  parent order equals the maximum child order, incremented by one if two
  or more children share that maximum.

\subsubsection{Horton Ratios}
Following the Horton--Strahler framework~\cite{horton1945}, three
  self-similarity ratios are computed between consecutive Strahler
  orders~$n$ and~$n\!+\!1$:
  \begin{equation}
    R_B = \frac{N_n}{N_{n+1}}\,,\quad
    R_L = \frac{\bar{L}_{n+1}}{\bar{L}_n}\,,\quad
    R_D = \frac{\bar{D}_{n+1}}{\bar{D}_n}\,,
    \label{eq:horton}
  \end{equation}
  where $N_n$ is the number of segments, $\bar{L}_n$ is the mean length,
  and $\bar{D}_n$ is the mean diameter at order~$n$.

\subsubsection{Tortuosity}

Vessel tortuosity~\cite{bullitt2003} is defined for each edge as the ratio
  of the traced path length to the Euclidean distance between its
  endpoints:
  \begin{equation}
    \tau = \frac{L_{\mathrm{path}}}{L_{\mathrm{Euclidean}}}\,,
    \label{eq:tortuosity}
  \end{equation}
  where $\tau = 1$ indicates a perfectly straight segment.

\subsubsection{Asymmetry Ratio}

    Symmetry at each bifurcation is quantified by
    the ratio of the two daughter radii~\cite{zamir1999}:
    \begin{equation}
      \delta = \frac{r_{\min}}{r_{\max}}\,,
      \label{eq:asymmetry}
    \end{equation}
    where $r_{\max}$ and $r_{\min}$ are the largest and
    second-largest daughter radii at a given junction (the overall largest-radius neighbor is taken as the parent). A value of $\delta = 1$ indicates a perfectly symmetric split, $\delta \to 0$ indicates strong asymmetry.

\section{Results}
\label{sec:results}

\subsection{Morphometric Results}
\label{ssec:morphometric_results}

The pipeline was executed with three configurations that differed only in the vesselness fusion stage: Frangi only, Sato only, and the weighted Frangi+Sato fusion with $w_{\mathrm{F}}=1.0$ and $w_{\mathrm{S}}=1.2$. All other parameters were held constant. Table~\ref{tab:morphometry} summarizes the resulting morphometric metrics, and Figs.~\ref{fig:fractal_sweep} and~\ref{fig:murray_sweep} show the underlying fractal and Murray's law fits.

\subsubsection{Graph Topology}
The fused configuration produced the largest graph, with 12\,929 nodes and 4\,520 bifurcations, compared to 8\,433 nodes and 3\,351 bifurcations for Frangi alone, and 10\,113 nodes and 3\,534 bifurcations for Sato alone. This corresponds to 53\% more nodes than Frangi and 28\% more than Sato.

\subsubsection{Strahler Ordering and Asymmetry Ratio}
The maximum Strahler order was 7 for both Frangi and the fused configuration, and 6 for Sato alone. Mean bifurcation asymmetry was $\delta~=~0.90~\pm~0.10$ (Frangi), $0.87~\pm~0.12$~(Sato), and $0.88~\pm~0.12$ (fusion), pooled across all bifurcations of both lungs.

\subsubsection{Fractal Dimension}
\begin{figure}
    \centering
    \begin{tikzpicture}
    \begin{loglogaxis}[
        xlabel={Box size $\epsilon$ (voxels)},
        ylabel={Box count $N(\epsilon)$},
        width=0.95\linewidth,
        height=7cm,
        grid=major,
        legend pos=south west,
        mark size=2.5pt,
        tick label style={font=\footnotesize},
        label style={font=\small},
        legend style={font=\footnotesize}
    ]

    \addplot[only marks, mark=triangle*, color=red] coordinates {
        (2, 87816) (4, 19715) (8, 3971) (16, 739) (32, 166) (64, 47)
    };
    \addlegendentry{Sato only ($D_f^{\mathrm{vol}} = 2.21 \pm 0.04$)}
    \addplot[no markers, thick, red, forget plot] coordinates {
        (2, 85281.43) (64, 39.98)
    };

    \addplot[only marks, mark=square*, color=blue] coordinates {
        (2, 61066) (4, 17851) (8, 3910) (16, 748) (32, 166) (64, 48)
    };
    \addlegendentry{Frangi only ($D_f^{\mathrm{vol}} = 2.12 \pm 0.05$)}
    \addplot[no markers, thick, blue, forget plot] coordinates {
        (2, 67386.87) (64, 43.62)
    };

    \addplot[only marks, mark=*, color=darkgreen] coordinates {
        (2, 103293) (4, 23193) (8, 4330) (16, 781) (32, 171) (64, 48)
    };
    \addlegendentry{Frangi$+$Sato ($D_f^{\mathrm{vol}} = 2.26 \pm 0.05$)}
    \addplot[no markers, thick, darkgreen, forget plot] coordinates {
        (2, 100793.55) (64, 40.19)
    };

    \end{loglogaxis}
    \end{tikzpicture}
    \caption{Log-log box-counting plot of the volumetric
    fractal dimension $D_f^{\mathrm{vol}}$ of the reconstructed
    vascular graph, obtained from three complete pipeline runs
    that differ only in the vesselness fusion weights. Markers
    show non-empty box counts $N(\epsilon)$ on the skeleton
    inflated by the per-node EDT radius. Lines are the linear regressions whose negated slopes give $D_f^{\mathrm{vol}}$.}
    \label{fig:fractal_sweep}
\end{figure}
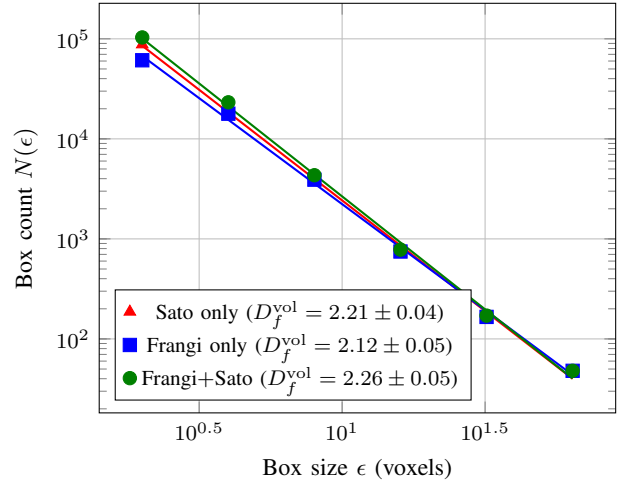

Log--log box-counting on the one-voxel skeleton yielded $D_f^{\mathrm{skel}} = 1.83~\pm~0.11$ (Frangi), $1.84~\pm~0.10$ (Sato), and $1.91~\pm~0.10$ (fusion). Inflating the skeleton by the per-node EDT radius and repeating the analysis gave volumetric dimensions of $D_f^{\mathrm{vol}} = 2.12~\pm~0.05$, $2.21~\pm~0.04$, and $2.26~\pm~0.05$, respectively (Fig.~\ref{fig:fractal_sweep}). The reported errors are linear-regression fit errors from the box-counting plot.

\subsubsection{Murray's Law}
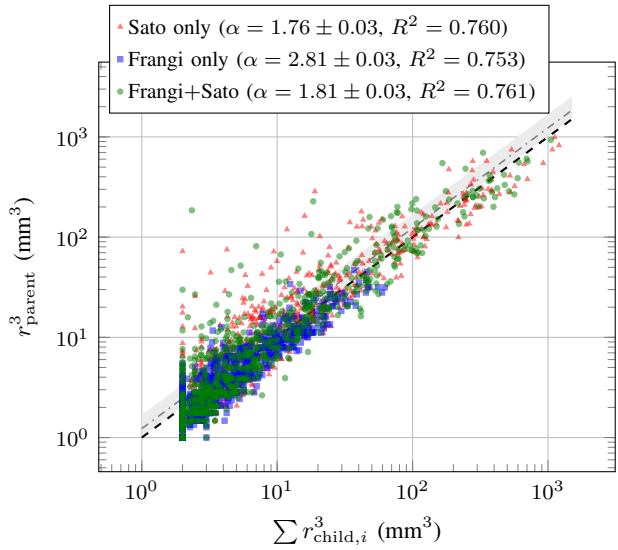
\begin{figure}
    \centering
    \begin{tikzpicture}
    \begin{loglogaxis}[
        xlabel={$\sum r_{\mathrm{child},i}^{3}$ (mm$^3$)},
        ylabel={$r_{\mathrm{parent}}^{3}$ (mm$^3$)},
        width=0.95\linewidth,
        height=7cm,
        grid=major,
        mark size=1pt,
        tick label style={font=\footnotesize},
        label style={font=\small},
legend style={
    at={(0.02,1.14)},
    anchor=north west,
    legend columns=1,
    font=\footnotesize,
    draw=black,
    fill=white,
    fill opacity=1,
    text opacity=1,
    cells={anchor=west},
},
    ]

    \addplot[no markers, thick, black, dashed, forget plot, domain=1.0000:1509.34] {x};

    \addplot[name path=murray_lo, no markers, draw=none, forget plot, domain=1.0000:1509.34] {1.0217*x};
    \addplot[name path=murray_hi, no markers, draw=none, forget plot, domain=1.0000:1509.34] {1.6869*x};
    \addplot[fill=black!8, forget plot] fill between[of=murray_lo and murray_hi];

    \addplot[no markers, semithick, black!50, dashdotted, forget plot, domain=1.0000:1509.34] {1.2300*x};

    \addplot[only marks, mark=triangle*, color=red, mark options={solid}, opacity=0.5] coordinates {
        (2.9279, 3.5820) (4.4124, 3.1856) (3.0162, 2.2515) (3.1028, 2.6400) (2.7589, 3.3750) (6.8454, 3.8931) (2.3439, 3.7914) (13.2262, 11.1803) (13.5638, 9.5016) (18.3830, 16.2666) (5.1891, 2.7267) (2.4740, 4.8115) (9.4013, 17.9443) (42.6945, 54.3585) (12.7657, 25.8584) (22.3607, 13.0581) (2.3439, 1.7589) (11.5700, 13.0581) (237.3111, 252.0318) (30.5353, 23.3207) (2.7437, 4.9749) (3.5178, 2.2515) (136.4288, 136.3989) (6.0227, 4.8115) (4.4644, 3.8931) (75.7706, 39.4325) (7.3569, 15.7350) (11.3231, 11.6279) (38.8693, 33.1406) (42.2185, 66.0610) (3.8371, 6.6811) (5.6569, 6.2864) (5.8183, 7.8102) (14.4913, 11.1803) (9.9659, 11.1125) (62.1470, 91.4419) (2.0000, 6.9663) (7.4745, 6.4976) (3.5840, 1.7589) (9.6832, 20.1405) (41.4144, 29.2505) (18.8240, 14.6969) (2.6879, 3.6619) (22.5922, 21.1936) (3.3216, 2.8284) (8.3500, 8.9819) (3.5178, 7.2098) (31.4203, 20.9366) (121.1054, 100.8348) (10.6835, 13.3353) (4.9862, 12.8586) (7.4161, 10.7799) (3.1028, 2.8284) (25.5701, 28.5585) (9.5390, 13.3353) (5.4684, 4.2490) (2.7589, 4.2512) (245.3078, 243.7923) (34.4463, 140.1812) (2.0000, 2.2008) (2.2217, 1.5840) (2.7589, 5.3579) (3.8371, 2.0782) (2.3786, 1.7589) (2.7589, 1.7589) (2.0000, 1.4740) (10.9789, 7.8102) (6.3713, 11.0114) (2.7589, 4.2613) (2.5840, 1.7589) (2.0000, 1.4740) (31.8210, 30.4850) (15.1304, 14.6969) (19.9060, 27.0263) (2.4740, 2.0729) (666.9080, 534.6940) (15.5259, 69.9646) (4.1724, 6.6811) (2.4740, 2.3704) (3.5523, 2.0782) (2.0000, 1.9252) (2.0000, 1.4740) (14.9049, 13.4532) (2.0000, 2.2211) (6.9194, 11.1202) (310.2315, 252.3527) (5.6401, 5.0012) (2.0000, 2.0729) (8.6049, 4.7672) (3.8371, 3.5129) (2.0000, 1.0000) (2.0000, 7.3199) (18.9662, 285.9464) (22.3607, 12.2818) (7.6424, 15.6250) (4.7183, 3.1856) (20.3907, 47.2834) (2.0000, 3.7149) (8.9022, 9.7304) (10.7574, 26.3808) (7.2273, 6.6811) (7.7862, 4.7672) (538.2991, 746.4499) (4.5873, 6.0798) (5.2800, 4.2256) (116.0065, 175.7000) (5.1319, 10.7515) (2.0000, 1.7589) (2.0000, 3.4850) (41.8874, 180.0447) (2.4740, 12.3107) (9.7934, 6.9754) (3.9354, 3.3215) (2.2217, 1.5840) (7.5538, 15.5067) (7.3452, 8.0000) (2.7589, 2.2430) (7.7452, 42.9268) (2.0000, 1.6602) (14.0009, 21.8043) (278.1752, 572.6033) (2.0000, 1.4740) (2.3439, 1.4740) (5.0263, 2.0782) (2.0000, 1.0000) (224.6906, 204.5146) (3.1028, 4.0622) (1206.3793, 824.0729) (3.3198, 2.0782) (183.8682, 277.8780) (50.4063, 97.0080) (2.5421, 5.3369) (4.1293, 3.3750) (24.2713, 35.4912) (6.4489, 9.3193) (2.4740, 2.3704) (31.6095, 21.0087) (2.0000, 1.7589) (2.7589, 1.7589) (7.4260, 15.6250) (3.2329, 2.8284) (392.3762, 447.3212) (8.6038, 7.8102) (4.6786, 5.1962) (5.8199, 41.7196) (10.7530, 9.4303) (2.9252, 3.3750) (12.5538, 25.0107) (2.9827, 1.7589) (4.9067, 6.0798) (3.4740, 1.4740) (14.3529, 13.4532) (2.4740, 1.5421) (20.6507, 101.7275) (12.9128, 15.7126) (81.2151, 54.3517) (6.5606, 14.0711) (19.1653, 20.6917) (11.1314, 10.6114) (6.7965, 3.8931) (2.0000, 2.7267) (4.7183, 2.8284) (10.3907, 12.7580) (5.8183, 5.1962) (5.3112, 2.8284) (17.1199, 17.2856) (28.5557, 41.0304) (36.8437, 28.3266) (13.1271, 12.8586) (14.0121, 15.7126) (28.6906, 19.7293) (23.5053, 19.2888) (3.3428, 2.0782) (6.9628, 17.9443) (5.3263, 5.7749) (22.7745, 18.7655) (6.2969, 12.8586) (2.0000, 1.7589) (8.8874, 17.9443) (83.6970, 73.0103) (2.0000, 1.4740) (5.7506, 6.4976) (2.0000, 2.3704) (287.8715, 194.8524) (3.1028, 1.7589) (8.3316, 29.6305) (40.1525, 44.1676) (218.4722, 145.3213) (25.3103, 38.3010) (43.1900, 148.9383) (2.0000, 1.6327) (36.1502, 25.9687) (6.2679, 15.9547) (2.0000, 1.3439) (2.0000, 1.7589) (31.0545, 26.1520) (2.0000, 1.7589) (15.8808, 20.1954) (4.8502, 3.9245) (5.4793, 6.0422) (3.8355, 4.1113) (4.1141, 28.9272) (3.8371, 2.6400) (14.3864, 16.6260) (2.0000, 1.4740) (3.2329, 3.7892) (87.6851, 127.4919) (2.9996, 4.2361) (2.4740, 1.4740) (27.2845, 55.5739) (102.8572, 83.8135) (22.8429, 13.1484) (7.6573, 6.4976) (149.8721, 136.3989) (3.9097, 8.4485) (261.9695, 136.3989) (5.6569, 4.9286) (3.1028, 7.1551) (3.7827, 13.2452) (2.0000, 1.0000) (28.8555, 37.1140) (15.2825, 14.4824) (2.0000, 2.4799) (4.1303, 2.5490) (131.3428, 134.4776) (7.1620, 13.1484) (2.4740, 1.9252) (2.9481, 2.3479) (11.1856, 16.6373) (8.7068, 4.2613) (5.0799, 3.8931) (42.7665, 44.1392) (5.3056, 10.1256) (12.3758, 7.1187) (3.1028, 3.4573) (4.1565, 2.3479) (4.1565, 2.1296) (21.8626, 11.0114) (15.5814, 10.0411) (12.8008, 10.1256) (2.7589, 2.5490) (2.0000, 1.7589) (16.5026, 11.1803) (59.5770, 152.2824) (2.0000, 1.0000) (9.4403, 71.8123) (7.0000, 9.5016) (12.1234, 9.0580) (2.0000, 3.9869) (7.3315, 6.0798) (3.2211, 44.2464) (5.4177, 8.9819) (4.9067, 4.9749) (567.2187, 321.9969) (249.5680, 133.1975) (21.0767, 22.4890) (3.8153, 3.7246) (22.2799, 19.7293) (120.2339, 168.2889) (160.8957, 215.9375) (22.1173, 19.7293) (4.5652, 9.3194) (2.0000, 1.0000) (2.0000, 1.4740) (101.6491, 96.2341) (78.6972, 92.7970) (2.0000, 4.2361) (2.0000, 2.3704) (2.0000, 4.2361) (40.4272, 57.7061) (10.3705, 9.5016) (12.7065, 11.1803) (14.8656, 11.1803) (2.0000, 1.0000) (14.0433, 11.1803) (3.5523, 5.7749) (17.2601, 43.2908) (8.6538, 11.1803) (3.2329, 2.1296) (2.0000, 1.4740) (3.0286, 2.4799) (7.9930, 22.8020) (8.9082, 6.6811) (2.8180, 2.8284) (4.3025, 5.4454) (2.3439, 2.5286) (4.6786, 12.7580) (4.4124, 7.5540) (27.6726, 22.4846) (18.1754, 15.7126) (10.2341, 9.5016) (3.5490, 6.3769) (26.9065, 43.1721) (3.5523, 2.6400) (5.7390, 4.9749) (3.8930, 4.4801) (13.4221, 31.0131) (3.2329, 1.7589) (2.4740, 25.7952) (3.8371, 6.0798) (3.2329, 1.7589) (7.4513, 16.2375) (30.3991, 27.9910) (13.9989, 14.6745) (45.2995, 32.5215) (2.6896, 1.7589) (9.4583, 6.8596) (3.5482, 2.8284) (6.3713, 8.0000) (103.8069, 71.0923) (2.4740, 2.1296) (11.7326, 11.0114) (2.8180, 3.5129) (2.4740, 2.6400) (33.4491, 78.1386) (21.7641, 21.0087) (2.7589, 5.5918) (28.9604, 38.7488) (3.6841, 2.5490) (101.9599, 72.8634) (127.0079, 95.5438) (2.4740, 1.4740) (2.8180, 1.4740) (2.0000, 1.7589) (2.6556, 4.2361) (78.2871, 105.3934) (176.8137, 125.0000) (20.1865, 28.2747) (17.1015, 12.8586) (4.3759, 8.8772) (77.1430, 68.0280) (21.5733, 32.5955) (4.3989, 2.8284) (24.3005, 19.4253) (26.9938, 23.4610) (4.2071, 5.0479) (23.0284, 24.3148) (2.9481, 3.1738) (45.6464, 42.2069) (3.9081, 2.8284) (4.6092, 3.1738) (2.4740, 4.5416) (57.1120, 42.2069) (5.8111, 24.0024) (2.5322, 7.3635) (36.8557, 33.1097) (3.6240, 5.3811) (15.9464, 22.4133) (8.4228, 9.1394) (2.0000, 1.4740) (4.3025, 2.8284) (11.5348, 9.5016) (4.2271, 2.8284) (2.0000, 3.7493) (3.8371, 2.0782) (4.9067, 6.9089) (2.0000, 2.1296) (4.6786, 3.8931) (3.3479, 5.0479) (7.2995, 20.1405) (10.6336, 6.0798) (2.2697, 1.4740) (3.2697, 1.7589) (2.0000, 4.2361) (7.9045, 7.8102) (3.8371, 2.2515) (16.4972, 14.6969) (2.0000, 20.2689) (28.2459, 26.8747) (8.8552, 18.3780) (3.3315, 3.1856) (3.2902, 4.1692) (11.8098, 119.2364) (3.6028, 3.5912) (11.3750, 8.0000) (4.4975, 7.8732) (4.0548, 6.9754) (3.1028, 2.8284) (5.2800, 8.7502) (120.3800, 99.1847) (3.1067, 3.1856) (4.3025, 6.8475) (12.3258, 6.0798) (185.2232, 474.0493) (5.9306, 6.0422) (7.0860, 4.9749) (38.8649, 32.4626) (1076.9853, 747.6645) (9.7154, 7.2228) (10.3907, 11.8561) (8.4910, 5.1962) (7.7346, 18.3780) (2.0000, 1.9462) (4.0245, 2.8284) (3.2329, 2.1112) (2.3439, 1.4740) (5.0873, 4.1692) (18.3353, 18.3766) (3.3992, 2.5490) (17.8614, 19.6649) (269.4882, 334.5464) (4.5498, 2.6400) (1118.8899, 988.7071) (282.2273, 373.1005) (2.0000, 1.9252) (2.7589, 1.7589) (14.1439, 28.9033) (5.5881, 2.6400) (49.3868, 127.4482) (139.0273, 131.9171) (24.9489, 50.6584) (5.5560, 3.9245) (25.5022, 16.4690) (4.4221, 9.0332) (27.5364, 31.3028) (2.3439, 2.3704) (8.0260, 14.3327) (2.4740, 2.3704) (180.8332, 177.9561) (5.4578, 13.8004) (123.3764, 80.5734) (3.5178, 5.4454) (2.0000, 3.5820) (4.7070, 2.6400) (2.0000, 1.0000) (19.3774, 29.5872) (2.9279, 2.3704) (57.2647, 71.4276) (20.6820, 15.4515) (352.7542, 575.2770) (23.8151, 43.4904) (8.3204, 15.9293) (4.3025, 3.8931) (6.0141, 3.7246) (18.4805, 16.8137) (83.7146, 160.7447) (361.4457, 300.4684) (102.8223, 115.9241) (4.3989, 3.8931) (116.5271, 120.1466) (6.1420, 6.0798) (2.8180, 1.7589) (4.2957, 16.1615) (18.1252, 41.5692) (2.4740, 2.7519) (3.5523, 2.0782) (2.0000, 6.5358) (13.1729, 11.1803) (13.9046, 45.0070) (24.6069, 31.5040) (6.2949, 3.9245) (3.6148, 3.9245) (35.1564, 28.8103) (3.4957, 1.7589) (3.4222, 7.5540) (2.0000, 8.7502) (10.0178, 9.2490) (2.0000, 1.3004) (6.4540, 8.9768) (9.5033, 8.9768) (3.4485, 6.1603) (11.2418, 11.1803) (3.0000, 2.7560) (3.2077, 7.5540) (5.9349, 3.8931) (16.5832, 41.2717) (8.1605, 7.8732) (4.5873, 3.5129) (15.9573, 162.5747) (9.9531, 57.5473) (2.7589, 2.0782) (7.8362, 9.7894) (4.0981, 3.8931) (12.1220, 7.8732) (2.3439, 1.9252) (2.3439, 2.5490) (76.9356, 94.5757) (326.6179, 305.7864) (2.0000, 1.0000) (2.7589, 1.7589) (705.4158, 378.6187) (4.3025, 2.8284) (2.0000, 1.3439) (4.2047, 6.4976) (2.4740, 3.3750) (523.4281, 410.5554) (2.0000, 4.4627) (2.6375, 1.7589) (2.0000, 2.1296) (14.1872, 11.1125) (9.2984, 9.5016) (7.8362, 5.1962) (4.2926, 3.8931) (2.0000, 72.1640) (5.1891, 3.2949) (2.0000, 2.5490) (18.6086, 16.2375) (5.8183, 3.8931) (381.0336, 374.7315) (18.5140, 17.8535) (20.7736, 19.0687) (6.7028, 6.5401) (4.1724, 2.8284) (2.7589, 3.7149) (6.8488, 18.6918) (79.3697, 69.5083) (2.0000, 3.0311) (2.0000, 2.7580) (6.9652, 4.9749) (4.8112, 4.6296) (2.7589, 2.5868) (5.9676, 8.8449) (49.8356, 43.2578) (2.0000, 3.7246) (3.8371, 3.8931) (3.2329, 6.0798) (4.1141, 6.6811) (3.4490, 2.8284) (5.2639, 3.4573) (18.3099, 14.2777) (2.0000, 1.7589) (11.3465, 31.5040) (4.5388, 3.3750) (17.0052, 12.8586) (99.1553, 107.0696) (3.3704, 2.4799) (3.4732, 5.6138) (6.9628, 6.0689) (16.4653, 20.7442) (3.8371, 3.1856) (6.5331, 6.4976) (11.1272, 8.0000) (4.2957, 11.8561) (237.0378, 271.1852) (45.6113, 36.4829) (4.4868, 5.9938) (2.4740, 2.8153) (11.1588, 18.2326) (19.0800, 27.9268) (21.6201, 17.9014) (6.0892, 17.1450) (3.0286, 3.8931) (4.5704, 3.4573) (8.1453, 11.0956) (138.8922, 161.2839) (3.5523, 2.0782) (268.4450, 169.0556) (6.5399, 16.2375) (6.8837, 4.9749) (5.7328, 17.1963) (8.6918, 11.0114) (70.2348, 67.2667) (3.5901, 2.0782) (215.3904, 146.0614) (5.5587, 24.6602) (82.9044, 71.6279) (23.7148, 46.1679) (56.4779, 139.0733) (2.0000, 3.3750) (13.7057, 17.9443) (2.4740, 1.7589) (15.0277, 19.2888) (2.4740, 1.7368) (34.2900, 43.4787) (2.0000, 12.8116) (2.0000, 1.0000) (3.5523, 7.5540) (275.2411, 463.9660) (2.3439, 3.1856) (4.7183, 3.5129) (17.5029, 16.4690) (2.0000, 1.0000) (38.2452, 38.7488) (56.5377, 74.9399) (9.0076, 49.5878) (4.1894, 7.5540) (32.6196, 41.3425) (2.4740, 1.4740) (2.3439, 1.4740) (2.0000, 2.1296) (18.2789, 14.2777) (2.5657, 2.5490) (77.7453, 76.2814) (2.7589, 9.5722) (2.7589, 1.9252) (3.4222, 4.8111) (4.5913, 2.8284) (4.5637, 2.8284) (65.9441, 74.0398) (24.9243, 52.3832) (4.4124, 2.8284) (3.1500, 2.5490) (4.9067, 5.4454) (6.5897, 3.2949) (3.5178, 3.1738) (6.5331, 8.7484) (83.8428, 85.7315) (9.0893, 5.1962) (72.2938, 91.7975) (2.0000, 1.0000) (3.2329, 3.8931) (3.6296, 3.8931) (2.3439, 1.7589) (24.9624, 12.8586) (5.4684, 2.8284) (3.3882, 4.2998) (9.2761, 8.5800) (4.7747, 4.5403) (4.0212, 3.1856) (2.5840, 1.7589) (2.3439, 3.6891) (42.4148, 127.8557) (3.2329, 2.8284) (6.8106, 15.2631) (2.7589, 1.7589) (3.8371, 3.8931) (4.1565, 2.0782) (3.8371, 2.6863) (4.1724, 3.8931) (13.0237, 23.4846) (2.0000, 1.0000) (2.0000, 1.9252) (9.5539, 6.0798) (2.0000, 1.7589) (2.4740, 1.7589) (14.2701, 21.0087) (3.9840, 4.9628) (2.0000, 2.8153) (2.7589, 4.4177) (3.9871, 4.7338) (2.4740, 1.7589) (12.5559, 7.8732) (6.5897, 6.4976) (2.7589, 9.0764) (6.7215, 3.8931) (6.4006, 15.1244) (3.8371, 3.8931) (4.3025, 3.1856) (3.4222, 3.1856) (3.2329, 2.0782) (3.5523, 2.8284) (13.2250, 16.2375) (9.5522, 4.9749) (37.0045, 53.4920) (12.2035, 38.6722) (5.9296, 9.5016) (15.7565, 30.2196) (8.4786, 6.8596) (2.0000, 3.3750) (2.0000, 1.7589) (22.4689, 17.9443) (8.4905, 17.4283) (104.5032, 104.9482) (6.5493, 15.6250) (9.0164, 9.1038) (6.4687, 3.4337) (5.2439, 4.8111) (3.5523, 3.1856) (10.4408, 9.5016) (73.9073, 55.8936) (250.3600, 233.8587) (3.2217, 1.7589) (61.3204, 38.7488) (17.7049, 11.1803) (11.8640, 24.7494) (7.8843, 11.0114) (53.9098, 85.0674) (3.2159, 6.2864) (4.2926, 2.8284) (3.2329, 2.3704) (309.9984, 198.1881) (3.4740, 3.0394) (3.8371, 2.8284) (8.8680, 8.0000) (3.7255, 3.8931) (921.1864, 547.5694) (11.4725, 8.0000) (5.5487, 3.8931) (23.7873, 65.3606) (56.8273, 60.7988) (2.0000, 1.7589) (2.7437, 5.9253) (15.5814, 11.1803) (9.0414, 18.9782) (4.3025, 6.6811) (3.2329, 3.2605) (436.6845, 268.6625) (5.6372, 22.1098) (6.5331, 20.3614) (2.0000, 4.2361) (2.7589, 2.6400) (5.0521, 5.3168) (3.3916, 2.0782) (24.3795, 79.6961) (249.7950, 275.2361) (4.2572, 75.1717) (2.0000, 17.6129) (2.0000, 1.7589) (74.3354, 128.6214) (246.0843, 241.0680) (53.1208, 116.1323) (34.6154, 24.7441) (106.7925, 123.0191) (108.8940, 62.7539) (322.9957, 202.5806) (382.3387, 295.6763) (2.4740, 4.1918) (241.9623, 210.1552) (25.8623, 21.2351) (3.0580, 1.7589) (3.2329, 1.7589) (82.4088, 41.0304) (3.4740, 1.7589) (2.7589, 8.0000) (17.4169, 26.7976) (38.7789, 126.9314) (5.2718, 15.6661) (22.8665, 16.2375) (5.1891, 2.7067) (292.2452, 206.4915) (2.0000, 1.0000) (12.4609, 9.7894) (70.9853, 88.5399) (3.3704, 4.9749) (16.5493, 27.5999) (218.2216, 99.6712) (3.2329, 2.7143) (4.8526, 15.8925) (7.0000, 14.0711) (3.5490, 3.0588) (2.9531, 2.5490) (7.8362, 19.6644) (47.7106, 57.7061) (7.3548, 9.5016) (75.9532, 111.5645) (179.1918, 220.6959) (2.4740, 2.1296) (10.3750, 16.6260) (6.0541, 32.5767) (13.8487, 27.8113) (52.5235, 69.2008) (5.1883, 5.7749) (11.8868, 17.0767) (22.5218, 32.6320) (4.2078, 3.7246) (386.8983, 285.0886) (31.1175, 27.7909) (4.3989, 3.5129) (5.2467, 3.6525) (16.1169, 22.6274) (5.8360, 2.8284) (6.3807, 4.1692) (2.3987, 2.6400) (6.5331, 8.6491) (5.7390, 4.2993) (20.6401, 51.9089) (14.5811, 16.2375) (3.3439, 1.7589) (2.0000, 3.5820) (8.7691, 17.1979) (10.7916, 6.0798) (2.7722, 1.7589) (2.0000, 4.4694) (4.6583, 14.0328) (2.0000, 1.0000) (3.4222, 2.0782) (527.8104, 280.6346) (28.7842, 29.2505) (244.3075, 228.8883) (2.4740, 3.3750) (331.6084, 262.4695) (7.2488, 6.4976) (13.4641, 14.6969) (3.2329, 2.0782) (3.5523, 2.7143) (3.0782, 2.5490) (2.2697, 3.3163) (7.4060, 9.1394) (3.7255, 2.8284) (3.1028, 1.9252) (88.3236, 74.2071) (5.6569, 3.4850) (48.2104, 31.1910) (13.4021, 16.5816) (2.6556, 2.5490) (2.9252, 4.5416) (2.7437, 2.6863) (64.0450, 86.0989) (3.4203, 6.4063) (6.7965, 2.8284) (2.9481, 2.8153) (3.2329, 2.7143) (2.0000, 2.0782) (17.8614, 18.2308) (2.0000, 1.8575) (4.2590, 5.9894) (4.4611, 6.6073) (2.0000, 1.0000) (2.0000, 1.3439) (2.0000, 1.3439)
    };
    \addlegendentry{Sato only ($\alpha = 1.76 \pm 0.03$, $R^2 = 0.760$)}

    \addplot[only marks, mark=square*, color=blue, mark options={solid}, opacity=0.5] coordinates {
        (2.0000, 1.4740) (3.3428, 2.9196) (3.3000, 3.5129) (2.0000, 1.0000) (2.3987, 1.8914) (8.9739, 11.1803) (10.8999, 8.0000) (8.2926, 8.0000) (8.6108, 6.6610) (3.2515, 2.2638) (2.0000, 1.0000) (6.3360, 4.5133) (4.1141, 3.4850) (2.0000, 1.4740) (3.5523, 6.3682) (2.0000, 1.7589) (2.0000, 1.0000) (2.0000, 1.0000) (2.0000, 1.0000) (2.7589, 1.9462) (4.0981, 5.0479) (2.0000, 1.7589) (2.0000, 1.0000) (2.0000, 1.1882) (3.5523, 2.6863) (2.0000, 1.2217) (3.4485, 6.9754) (3.2329, 3.4573) (2.0000, 1.7589) (2.3439, 2.5490) (4.0866, 1.4740) (3.6046, 5.8782) (2.0000, 1.7589) (4.9067, 3.5129) (2.4740, 1.7589) (2.0000, 1.0000) (8.3545, 10.0680) (2.0000, 1.0000) (8.1802, 8.1894) (2.2697, 2.7139) (2.0000, 2.1112) (3.0286, 2.8284) (8.7091, 5.1962) (2.0000, 1.0000) (9.8869, 10.0411) (2.0000, 1.3987) (3.2902, 3.1583) (2.0000, 2.5536) (18.3465, 11.1803) (8.1605, 5.8782) (12.2285, 7.8102) (18.5731, 11.1803) (10.6693, 8.4954) (5.8138, 3.8931) (2.0000, 2.6863) (3.6205, 2.8284) (2.0000, 1.0000) (3.0782, 2.1296) (47.9681, 31.6228) (2.0000, 1.0000) (2.7589, 2.0782) (14.9247, 12.2818) (10.0958, 8.6796) (3.0842, 1.7589) (12.3087, 19.7293) (46.5948, 31.6228) (2.0000, 1.2697) (3.5697, 5.0479) (12.4279, 10.0411) (55.6757, 31.6228) (4.7183, 2.8284) (3.0908, 2.8205) (27.9011, 24.0279) (5.4684, 3.3325) (4.7747, 3.8931) (2.2697, 2.3704) (2.0000, 1.0000) (3.0000, 1.2697) (2.0000, 3.8628) (3.3479, 4.9749) (2.0000, 3.3750) (8.8376, 6.2864) (12.6081, 12.8586) (7.5198, 7.8102) (3.2137, 7.8102) (2.7589, 1.9462) (3.4203, 4.2490) (3.4203, 3.8931) (6.2404, 11.1803) (4.0311, 3.9337) (17.2215, 14.6969) (2.0000, 1.3439) (7.7867, 13.4532) (2.0000, 1.2697) (2.4740, 1.9252) (3.0000, 4.0622) (3.5716, 5.8956) (9.6054, 9.5016) (2.8180, 2.4130) (2.0000, 1.3439) (2.8180, 2.0782) (4.8439, 3.0557) (4.4690, 3.2272) (2.0000, 2.3394) (2.6958, 1.4740) (25.5223, 20.4291) (6.1383, 6.0422) (5.1735, 4.0235) (31.5493, 22.6274) (2.0000, 1.0000) (22.0634, 22.6274) (3.8371, 3.8931) (4.8112, 4.9749) (2.3439, 2.0782) (6.7722, 6.7132) (3.6046, 2.8284) (2.9481, 2.8284) (5.6464, 6.2281) (2.3439, 1.4740) (2.4332, 1.3439) (2.0000, 1.0000) (2.0000, 1.4740) (2.9481, 2.0782) (3.3479, 2.1464) (2.1882, 1.4740) (2.4100, 1.7589) (2.7589, 1.7589) (2.0000, 1.7589) (4.1898, 3.3325) (2.8180, 2.8284) (5.9298, 5.0847) (4.8439, 4.9662) (2.1635, 1.7589) (5.4621, 4.2998) (2.0000, 1.7589) (5.7026, 4.0954) (2.3439, 1.6896) (5.2801, 2.4282) (15.2665, 15.7126) (10.1632, 10.5732) (2.0000, 1.0000) (17.2586, 21.9239) (14.6704, 12.8586) (4.7696, 6.0422) (8.6908, 7.5540) (24.5166, 31.7071) (4.1201, 8.0000) (4.3025, 3.7493) (2.0000, 1.4740) (3.4222, 2.8284) (3.7752, 2.3599) (5.0326, 5.7749) (3.5149, 5.8227) (2.0000, 1.0000) (11.5682, 16.2375) (5.9349, 5.7024) (2.0000, 1.9252) (6.1865, 5.4622) (3.8371, 3.5129) (16.5338, 14.2777) (6.4751, 4.1019) (10.5621, 8.7293) (2.0000, 1.0000) (23.4621, 13.5650) (2.0000, 1.3439) (2.0000, 1.0000) (2.0000, 1.5840) (2.3439, 1.4740) (5.1852, 3.3325) (3.0286, 2.2515) (22.0788, 12.2818) (9.5931, 11.0675) (15.4164, 15.7126) (14.4193, 11.9999) (2.7437, 1.5840) (24.4982, 26.9164) (7.8769, 4.9749) (11.2450, 7.8102) (24.7513, 27.0000) (6.6089, 7.1897) (4.4975, 5.8227) (31.8533, 25.8584) (2.0000, 1.4740) (5.8190, 3.8931) (2.0000, 1.0000) (2.0000, 2.1761) (6.1446, 4.9055) (3.3428, 11.6930) (8.1929, 9.0351) (3.5523, 2.8284) (12.9344, 15.4544) (5.5827, 7.8102) (4.3025, 3.8931) (33.1608, 20.6762) (2.0000, 1.0000) (2.0000, 2.8852) (11.6881, 9.5016) (13.8978, 7.3018) (16.5405, 20.6762) (2.0000, 2.0782) (3.8371, 2.8284) (2.3439, 2.0782) (3.6501, 6.0798) (2.0000, 1.2217) (4.4124, 3.8931) (2.0000, 1.0000) (3.4993, 2.8284) (2.9481, 3.9602) (3.3000, 8.9531) (23.1915, 19.4253) (25.8947, 20.4291) (16.6369, 12.8586) (2.0000, 1.7589) (6.2679, 8.1105) (8.8580, 7.2300) (3.8355, 3.5437) (7.6604, 9.7304) (20.6585, 20.4291) (4.0104, 7.6830) (4.2329, 1.7589) (23.3221, 23.6725) (19.5112, 16.8061) (10.0596, 6.8272) (7.4060, 4.3528) (2.7589, 2.2515) (3.6622, 4.4283) (5.2800, 6.4976) (2.0000, 1.0000) (2.7589, 4.1234) (2.0000, 1.0000) (6.5331, 5.2577) (2.6879, 1.7589) (2.4740, 1.4740) (4.0548, 6.4976) (10.9855, 8.0000) (3.5523, 2.0782) (4.3297, 3.8931) (2.0000, 1.7589) (3.1296, 3.6279) (3.0286, 4.9749) (2.5873, 1.7589) (2.0000, 3.2488) (12.4833, 34.0042) (2.0000, 1.6327) (4.5873, 4.2998) (6.5294, 6.5251) (5.6650, 2.9812) (4.5124, 3.5129) (3.5471, 3.0937) (21.1937, 11.0956) (2.0000, 1.4740) (3.1342, 4.2998) (3.1808, 1.7589) (3.0580, 4.0622) (15.8320, 11.1803) (2.0000, 3.0937) (7.0430, 6.2864) (16.9232, 10.0411) (4.1870, 2.8284) (15.4746, 11.1803) (5.0468, 6.4976) (7.8490, 9.5016) (19.4054, 24.7494) (2.0000, 1.4740) (4.8152, 4.9749) (15.2528, 12.2818) (9.7738, 9.3948) (10.3220, 6.9754) (5.3448, 6.9754) (4.1565, 2.8284) (5.2787, 6.6253) (2.7589, 2.6400) (3.7051, 4.2998) (4.3933, 2.8284) (2.1635, 1.6602) (3.5523, 2.2515) (38.0017, 20.6945) (2.0000, 1.0000) (14.5318, 7.4447) (2.0000, 1.3987) (2.6684, 2.6863) (5.6655, 8.0000) (8.0162, 8.0000) (12.7787, 11.1803) (3.8371, 2.8284) (4.7463, 6.4976) (2.3439, 2.6034) (9.2634, 13.1327) (7.8917, 7.8732) (10.3923, 6.4976) (3.2217, 1.7589) (2.0000, 1.5840) (2.0000, 1.0000) (2.0000, 1.4740) (2.0000, 1.7589) (6.7693, 7.8102) (3.5523, 3.0937) (2.0000, 1.0000) (3.9650, 6.0422) (2.0000, 1.3439) (4.3487, 6.4976) (3.2329, 2.8284) (4.7475, 3.8931) (3.7656, 5.1962) (5.8306, 5.1962) (22.9406, 16.1038) (2.0000, 2.1296) (4.9067, 2.8284) (2.0000, 1.7589) (4.1565, 4.2490) (2.0000, 1.7589) (2.0000, 1.7589) (5.3479, 7.8102) (6.0723, 8.3729) (15.3094, 11.1803) (2.9279, 3.2272) (3.2329, 2.3599) (4.9505, 5.1962) (18.0893, 12.2818) (2.0000, 1.5840) (9.5056, 9.1038) (2.6136, 1.4740) (12.7967, 11.9999) (3.2515, 2.4880) (3.8478, 2.7331) (2.6549, 1.7589) (2.0000, 3.9398) (3.7051, 3.8931) (2.1445, 3.8680) (2.1635, 2.3479) (2.9471, 4.1417) (2.0000, 1.4740) (2.2217, 2.4799) (3.5178, 1.7589) (2.7589, 2.2515) (2.0000, 2.1296) (2.2697, 2.2515) (16.5087, 11.1803) (4.1724, 2.8284) (3.4222, 2.8284) (5.8673, 3.8931) (5.3671, 7.6639) (2.0000, 1.0000) (4.4382, 3.5129) (2.9481, 2.8284) (2.4740, 1.7589) (3.5815, 3.8931) (4.0548, 3.8931) (18.0443, 11.1803) (2.3439, 1.7589) (11.6070, 11.1803) (2.0000, 1.0000) (2.4740, 2.4880) (2.8180, 3.5212) (4.5090, 3.8931) (2.6684, 1.4740) (2.3439, 1.4740) (3.4222, 4.2998) (62.2833, 31.5040) (2.0000, 1.4740) (2.5840, 1.8575) (3.5523, 2.0782) (13.5483, 11.9999) (2.3439, 1.4740) (7.6291, 5.7749) (8.5564, 7.8732) (2.0000, 1.3439) (7.1101, 6.4976) (8.6272, 9.3948) (5.1891, 3.8931) (2.0000, 1.2217) (40.0663, 34.6537) (3.0000, 1.0000) (2.8180, 1.4740) (2.0000, 1.1882) (13.1354, 8.0000) (2.0000, 1.4740) (3.0000, 2.0782) (5.6566, 4.9749) (3.2329, 2.8284) (3.4222, 5.5048) (2.0000, 3.4027) (5.5224, 7.8102) (2.3439, 1.7589) (2.8180, 1.9462) (10.4416, 7.8102) (2.5421, 1.7589) (2.0000, 1.3439) (2.2697, 1.4740) (2.0000, 2.0175) (3.7051, 2.0782) (2.7437, 1.5840) (2.0000, 1.4740) (11.5479, 15.7126) (6.4010, 4.3636) (13.3906, 11.0675) (14.8527, 29.2505) (4.7013, 4.9749) (2.4740, 1.7589) (10.5375, 8.7293) (3.8700, 12.5970) (5.8314, 5.4665) (2.0000, 1.7589) (5.5912, 4.2998) (2.5322, 3.8931) (3.6296, 2.8284) (2.5840, 1.9252) (2.9462, 2.6034) (4.0718, 2.8284) (3.2601, 2.7805) (10.1632, 11.1803) (2.0000, 1.9462) (3.8371, 3.1583) (2.1882, 1.7589) (4.5652, 3.8931) (5.4007, 6.8596) (16.4673, 11.1803) (2.0000, 1.7589) (8.9249, 11.9999) (9.8869, 9.8229) (15.6087, 12.8586) (3.2329, 1.7589) (4.9067, 2.8284) (2.0000, 1.5840) (13.3998, 9.1038) (5.9024, 4.7376) (2.0000, 1.3439) (5.4684, 4.7376) (4.3888, 2.8284) (59.5413, 28.0459) (2.0000, 4.1918) (4.3371, 2.8284) (7.8362, 17.2615) (4.8314, 7.6639) (2.2697, 2.6034) (2.0000, 1.7589) (13.3435, 8.0000) (2.0000, 4.5306) (2.3439, 1.4740) (4.1724, 2.8284) (5.6569, 3.3325) (46.0655, 44.9134) (30.2567, 27.8858) (37.0451, 46.8722) (4.2930, 2.6863) (13.9566, 10.2380) (12.1925, 14.6969) (22.1640, 14.6969) (5.0873, 3.4556) (7.4985, 10.2380) (16.3814, 14.6969) (4.7183, 2.8284) (3.4222, 4.9749) (2.0000, 1.7589) (2.4740, 2.0782) (3.8371, 2.0782) (6.7029, 6.4976) (2.5840, 1.6556) (2.0000, 1.9462) (9.4403, 11.0675) (3.9951, 8.8449) (2.0000, 1.4740) (6.1873, 4.9749) (7.3720, 7.8102) (15.5919, 11.8181) (3.0000, 3.3163) (2.0000, 1.0000) (10.2961, 12.8586) (4.1141, 4.4119) (4.8272, 6.4126) (6.5706, 6.0798) (4.9861, 6.0516) (2.0000, 1.7589) (2.9481, 3.0937) (9.7600, 8.0000) (2.0000, 1.0000) (2.9481, 2.6400) (34.8577, 22.6274) (5.2107, 4.2998) (45.4411, 37.3846) (2.6958, 1.7589) (5.6309, 6.4976) (3.2329, 3.8931) (3.5810, 2.8284) (25.0741, 24.7494) (18.6057, 23.5128) (2.0000, 1.0000) (11.1235, 15.1284) (4.9631, 3.5129) (4.5873, 3.5129) (2.6136, 3.7246) (9.6334, 6.6749) (3.6622, 4.9749) (45.2548, 27.9910) (2.7589, 2.6400) (4.2957, 6.4976) (2.3987, 1.6556) (3.1067, 3.9536) (2.0000, 1.0000) (2.8180, 4.2998) (3.2329, 2.8284) (2.0000, 1.0000) (2.0000, 1.0000) (5.9349, 7.8102) (7.6027, 8.0000) (2.0000, 5.1212) (2.0000, 1.9345) (3.0000, 1.5840) (5.6425, 6.4976) (2.8180, 2.8284) (3.8371, 2.6863) (2.0000, 1.7589) (9.6832, 9.5016) (2.8180, 1.5840) (2.0000, 1.0000) (2.8180, 1.5840) (12.9314, 9.5016) (5.6569, 2.8284) (2.0000, 1.0000) (9.0970, 5.9174) (3.5523, 2.8284) (3.0314, 2.8284) (2.0000, 1.0000) (2.0000, 2.0782) (4.9067, 5.6219) (13.5327, 12.8586) (2.7764, 1.7589) (2.0000, 1.0000) (2.5394, 2.3479) (3.4222, 4.1692) (3.2329, 3.3325) (2.2697, 1.2697) (12.4646, 8.8449) (4.1565, 2.8284) (4.8079, 2.6400) (12.7768, 10.3310) (9.2117, 6.8272) (3.5523, 2.4799) (6.5331, 3.8931) (2.7437, 1.7589) (11.7017, 7.5540) (7.2558, 4.0071) (2.0000, 1.0000) (2.0000, 1.0000) (12.9749, 8.0000) (2.0000, 2.6400) (10.4799, 9.5016) (5.3146, 7.8102) (15.7175, 10.1988) (2.0000, 1.7589) (2.0000, 1.7589) (3.5178, 3.9239) (2.9481, 1.5840) (2.0000, 1.0000) (2.4740, 1.7589) (2.0000, 1.0000) (2.2697, 1.4740) (2.2697, 1.7589) (6.1367, 8.1151) (9.3303, 9.5016) (4.3062, 2.3599) (2.8180, 2.8284) (38.7286, 26.8747) (9.2196, 6.4976) (5.6569, 3.8012) (16.1692, 22.4133) (4.0178, 2.8284) (2.0000, 1.3439) (2.0000, 1.7589) (7.0498, 3.2560) (3.7958, 3.8931) (4.1565, 3.8931) (6.1529, 4.2998) (3.1028, 2.2589) (4.1972, 5.8616) (6.4753, 8.7293) (2.2697, 1.7589) (2.8180, 2.2589) (13.0843, 12.8586) (2.7589, 4.5416) (2.9481, 2.8284) (2.4740, 2.0782) (2.0000, 5.5089) (3.2329, 3.7246) (2.9481, 2.8284) (2.9806, 1.7589) (5.0388, 2.8284) (4.8630, 4.1932) (3.1028, 2.5101) (19.7464, 14.6969) (3.8842, 3.8931) (2.7437, 2.2515) (5.3419, 4.6448) (4.4080, 3.1097) (6.3713, 4.9749) (7.7562, 4.1932) (2.8180, 1.5840) (11.2379, 11.0675) (9.2030, 8.0000) (11.1856, 8.6283) (9.1010, 7.8479) (2.0000, 1.7589) (2.7589, 1.7589) (2.7589, 1.7589) (3.0000, 2.5376) (2.0000, 1.0000) (4.6044, 2.3599) (4.1565, 2.8284) (3.1028, 2.8284) (6.8704, 4.6507) (2.0000, 1.3023) (2.0000, 1.3439) (4.0502, 4.3047) (3.4740, 2.2515) (2.4740, 5.1025) (4.2273, 3.3325) (4.1724, 4.1692) (9.3300, 7.8732) (2.2217, 1.3439) (2.0000, 1.7589) (2.9481, 2.0782) (12.8713, 9.0580) (6.9976, 4.5133) (2.7437, 1.4740) (2.0000, 2.2515) (3.4222, 2.6863) (15.4872, 11.1803) (2.7589, 3.1738) (2.9279, 3.1097) (2.0000, 1.4740) (6.2813, 5.0479) (13.2262, 11.1803) (16.6083, 10.0411) (3.9097, 3.4885) (2.0000, 1.2697) (7.2794, 7.8102) (6.0227, 7.2135) (2.4740, 2.5490) (29.7687, 18.3780) (2.0000, 2.6400) (2.3992, 2.0782) (2.7589, 2.0782) (21.8128, 16.2375) (6.1366, 5.5302) (3.7255, 3.0557) (5.6534, 8.0000) (6.2375, 4.0257) (27.2599, 22.6274) (22.1968, 16.2375) (10.2492, 9.5016) (2.0000, 2.2589) (3.2329, 5.4525) (2.0000, 1.7589) (2.0000, 1.7589) (10.3600, 14.6969) (5.1679, 3.8012) (3.2329, 3.8931) (8.0173, 15.7126) (3.1586, 3.4090) (8.6424, 9.2163) (2.9279, 3.7914) (4.1603, 2.8284) (2.0000, 1.0000) (2.7589, 1.7589) (2.0000, 1.0000) (3.8371, 3.1512) (2.9481, 2.8284) (4.1565, 2.0782) (2.0000, 1.4740) (2.0000, 2.6863) (3.5178, 6.0798) (3.2329, 2.8284) (16.7619, 10.7435) (4.1724, 5.7749) (2.4740, 1.5840) (4.3025, 3.5129) (2.0000, 1.4740) (2.0000, 1.4740) (2.4740, 1.7589) (4.2271, 2.8284) (7.0669, 10.6823) (3.2329, 2.8284) (13.2262, 11.1803) (3.2329, 5.6743) (9.4089, 8.0000) (21.8110, 13.1629) (18.0977, 11.1803) (2.7589, 2.2515) (2.0000, 1.4740) (15.9022, 10.4870) (4.4124, 2.8284) (2.0000, 1.4740) (2.4740, 1.7280) (8.1605, 6.4976) (2.8180, 1.4740) (8.7045, 11.1803) (2.0000, 1.7589) (13.2732, 16.2375) (2.0000, 2.8186) (4.4124, 4.1280) (4.4960, 4.9749) (2.0000, 2.9631) (14.9306, 8.8449) (4.2590, 7.8102) (6.1609, 3.8012) (7.2056, 6.9754) (24.5536, 13.4557) (7.1642, 3.8931) (2.0000, 1.7589) (5.6432, 6.9089) (4.9455, 3.8931) (5.1756, 4.1692) (11.0970, 9.5016) (4.4743, 6.8596) (2.0000, 1.0000) (5.8673, 4.1692) (9.6838, 10.4870) (4.4872, 6.4976) (5.4162, 3.8931) (2.8180, 1.4740) (5.7697, 6.1210) (4.4019, 2.4880) (2.0000, 1.5840) (2.2697, 1.3439) (5.3661, 3.7682) (5.2640, 4.4283) (9.0321, 7.8732) (3.2329, 2.8284) (6.5331, 3.8931) (4.4124, 2.9319) (13.8397, 12.2818) (3.3428, 2.8284) (2.2697, 1.4740) (5.0817, 4.4119) (4.1565, 2.4880) (6.4743, 5.8227) (2.0000, 1.3439) (3.4222, 2.8284) (6.7634, 2.8284) (4.9999, 2.8284) (2.4740, 2.0782) (3.4916, 5.0479) (2.8180, 2.8284) (8.4424, 4.9749) (2.7589, 1.9462) (4.7696, 4.2342) (2.0000, 1.7589) (2.0000, 3.4573) (28.2945, 25.0170) (2.9806, 2.8284) (4.0502, 4.7376) (2.0000, 1.0000) (3.7039, 3.6619) (19.6996, 18.2117) (4.4547, 2.0782) (2.0000, 1.4944) (2.0000, 1.4740) (2.0000, 1.0000) (26.5013, 24.0279) (2.4740, 1.9462) (5.5781, 4.2560) (4.9848, 2.8284) (4.5098, 6.0422) (6.2595, 4.2998) (2.0000, 1.0000) (17.2272, 12.8586) (11.9144, 15.0415) (14.7800, 11.1803) (2.0000, 2.0403) (3.0000, 1.3439) (2.7437, 2.8284) (8.2590, 5.4454) (4.9067, 5.4454) (2.0000, 2.0782) (2.0000, 1.7589) (5.4107, 3.9337) (2.8536, 4.6296) (11.2482, 9.7607) (3.0023, 2.8284) (20.0252, 12.5104) (4.7474, 5.1962) (3.5954, 2.8284) (5.8052, 4.4210) (4.0981, 2.8284) (2.2697, 2.0782) (2.2697, 1.7589) (2.6879, 1.4740) (2.0000, 2.0782) (5.6337, 5.8227) (2.0000, 2.2515) (3.5728, 2.8284) (8.1486, 6.9754) (8.4363, 7.8102) (2.5075, 3.8931) (3.5817, 3.8931) (2.0000, 1.3439) (2.5394, 1.7589) (2.0000, 1.1635) (2.6958, 2.3260) (2.4435, 2.8284)
    };
    \addlegendentry{Frangi only ($\alpha = 2.81 \pm 0.03$, $R^2 = 0.753$)}

    \addplot[only marks, mark=*, color=darkgreen, mark options={solid}, opacity=0.5] coordinates {
        (16.8266, 20.1774) (6.5645, 4.7229) (6.5331, 5.1962) (7.1101, 8.0000) (7.7381, 2.6400) (3.6223, 2.8284) (22.3603, 15.2605) (2.7589, 2.2515) (6.0227, 3.8931) (3.0000, 1.2906) (13.6897, 8.9819) (5.6425, 11.1966) (3.1028, 2.2515) (4.2957, 3.8931) (2.7589, 1.8575) (24.4247, 27.0000) (17.2064, 16.2375) (4.8962, 2.8284) (2.4740, 2.1761) (9.5667, 16.2375) (10.4221, 21.6767) (2.0000, 1.0000) (11.6045, 9.8136) (17.2166, 26.9092) (2.0000, 1.7589) (2.0000, 1.6896) (17.2492, 11.1803) (2.0000, 4.2993) (3.8371, 6.6811) (12.8868, 9.5016) (2.0000, 1.2217) (3.3439, 2.0008) (2.0000, 1.0000) (3.5178, 2.8284) (17.2256, 20.2291) (106.1864, 76.3675) (2.8180, 1.7117) (365.9136, 363.5311) (3.2329, 2.0782) (2.0000, 1.0000) (2.0000, 2.6400) (278.0873, 289.7074) (6.0634, 4.4119) (3.0580, 2.8284) (2.0000, 1.4740) (2.0000, 2.6400) (3.4150, 2.8284) (2.0000, 1.9252) (21.3511, 20.2291) (2.7589, 2.3704) (3.5523, 2.8284) (11.5567, 23.8319) (4.0981, 2.8284) (46.0576, 29.2505) (7.0000, 81.2656) (2.0000, 1.0000) (3.8371, 3.1856) (4.5652, 3.8931) (38.1734, 22.3423) (3.5523, 3.8931) (2.0000, 1.0000) (17.2977, 17.9443) (2.3439, 2.3599) (4.7696, 7.8102) (6.4735, 6.4976) (2.0000, 7.6412) (4.3992, 2.5490) (31.8531, 45.0573) (2.0000, 1.7589) (405.9098, 250.4250) (8.0825, 6.4976) (4.7070, 2.8284) (52.6984, 57.7061) (2.0000, 1.7589) (2.3439, 1.5840) (3.4222, 7.2098) (2.0000, 1.0000) (3.7051, 2.8284) (3.4222, 2.4341) (55.9834, 26.9164) (3.8371, 2.8284) (2.0000, 5.5710) (11.9979, 7.8102) (2.7589, 1.7589) (2.0000, 2.1462) (7.3548, 9.5016) (5.3671, 9.6915) (20.9847, 55.6115) (4.5581, 2.8284) (14.3335, 25.8584) (4.2956, 3.3750) (16.8248, 13.2023) (236.0878, 177.5798) (21.7512, 13.4532) (4.5612, 8.7484) (16.0828, 18.9785) (3.6279, 4.0572) (2.5067, 1.7589) (2.0000, 2.5490) (2.7589, 1.7589) (5.0045, 10.2612) (32.8150, 48.9601) (8.3669, 6.0798) (2.6958, 2.5490) (3.0782, 2.5490) (7.5282, 4.6296) (2.5067, 1.7589) (3.6960, 3.1738) (8.7777, 11.1803) (2.7589, 2.1296) (2.0000, 1.2906) (4.4743, 3.2949) (16.8930, 14.6969) (2.9252, 1.9252) (14.7617, 14.6969) (21.0417, 19.7293) (3.5523, 2.0782) (3.8371, 2.8284) (6.7823, 5.5928) (2.1882, 2.2513) (2.9531, 2.3704) (3.4222, 2.8284) (2.9471, 1.9383) (2.6443, 1.4740) (23.0113, 44.0791) (39.1396, 57.7061) (3.5523, 2.8284) (2.4740, 2.3704) (166.0766, 547.7310) (2.3439, 185.2384) (3.0000, 1.7589) (6.7823, 5.8782) (77.8243, 174.7604) (18.8846, 29.2505) (8.0781, 6.4976) (2.4740, 2.8153) (2.2697, 1.3439) (45.7777, 38.7488) (2.7589, 3.0937) (4.7407, 3.1856) (3.1028, 2.8284) (10.6255, 67.1426) (1051.3947, 933.0254) (257.5625, 267.3305) (4.2469, 10.3588) (2.9481, 3.0937) (245.9744, 481.3256) (4.2957, 9.1394) (2.2697, 2.2430) (5.8393, 7.0141) (8.0712, 11.8561) (12.7120, 17.7689) (3.6622, 3.7246) (12.7525, 15.7126) (13.7377, 9.5016) (2.4740, 2.4079) (2.3439, 4.2361) (12.5023, 7.8102) (2.0000, 1.5840) (5.4245, 8.8103) (4.1141, 3.8931) (2.4740, 26.8999) (3.5178, 1.7589) (47.1587, 121.9817) (4.9869, 3.8931) (52.8133, 69.9186) (43.9747, 41.1339) (8.8994, 8.0000) (3.0580, 1.7589) (4.6574, 4.8111) (96.9028, 74.4695) (6.4735, 9.2281) (2.7589, 2.5490) (4.3025, 17.9722) (3.4408, 2.5490) (2.4740, 4.2361) (4.5873, 4.9749) (5.2718, 9.5459) (2.9471, 3.4885) (8.8316, 16.2375) (2.3439, 5.1524) (8.4327, 11.1803) (121.2832, 117.5755) (58.3556, 66.4399) (9.8585, 7.8102) (2.4740, 2.8153) (2.0000, 1.7589) (6.5399, 5.3168) (2.0000, 1.0000) (2.7589, 2.5490) (2.2697, 6.1317) (79.4349, 74.1706) (6.9357, 7.6000) (2.0000, 1.0000) (22.2843, 20.2291) (2.6896, 1.7589) (4.2283, 3.8931) (6.7338, 12.9998) (7.7015, 5.1962) (2.7437, 1.7589) (124.5495, 102.8689) (95.4573, 82.5891) (108.8817, 117.2996) (3.5523, 21.1548) (4.5652, 2.7143) (3.9840, 2.8284) (3.8371, 2.8284) (3.8371, 4.1692) (2.8180, 1.7589) (2.0000, 1.7589) (31.8219, 27.3659) (2.0000, 1.5421) (4.9580, 3.3325) (10.8899, 18.3780) (2.0000, 1.7589) (8.5811, 5.8782) (2.0000, 4.7672) (3.8371, 6.6811) (10.5855, 11.8561) (5.3263, 3.9245) (2.9252, 4.2361) (23.9809, 19.0017) (6.4313, 8.4437) (2.0000, 1.4740) (2.0000, 1.4740) (2.7589, 2.6400) (4.1565, 3.7246) (72.3021, 79.1998) (5.9743, 4.9798) (4.2920, 7.4316) (2.0000, 4.8510) (17.8723, 26.8747) (3.0000, 1.4740) (2.0000, 1.3439) (2.0000, 1.4740) (2.0000, 1.0000) (3.5523, 2.0782) (4.4197, 2.5490) (12.8421, 27.9157) (3.5178, 2.5490) (11.8354, 13.4557) (2.0000, 1.0000) (30.9805, 24.7494) (20.3804, 45.2854) (3.5523, 2.6400) (2.0000, 2.1296) (33.5693, 31.5040) (16.4185, 22.2686) (2.0000, 1.4740) (52.6038, 46.6471) (13.9989, 11.8618) (2.0000, 1.9462) (5.3671, 4.9749) (2.0000, 1.4740) (11.3928, 15.4515) (8.9724, 9.5016) (2.0000, 1.0000) (2.8180, 2.6400) (2.7589, 2.0782) (17.0225, 27.1514) (2.9806, 4.9749) (2.0000, 1.7589) (17.2601, 26.4162) (6.8436, 6.0798) (220.3365, 135.4536) (2.1173, 1.4740) (153.8375, 151.9902) (3.8371, 2.8284) (2.0000, 1.7589) (14.4629, 10.0411) (150.3095, 127.3934) (15.9914, 24.9641) (9.0638, 8.0000) (596.5381, 387.7161) (2.0000, 1.0000) (3.4740, 2.4799) (2.0000, 1.0000) (27.2252, 27.0000) (9.4996, 7.8102) (17.9555, 20.2291) (6.9810, 9.7607) (2.7589, 3.7892) (77.2064, 200.7010) (3.2329, 2.1296) (2.4740, 1.7589) (18.2789, 24.9493) (6.5554, 5.0479) (2.9766, 1.8575) (3.2329, 1.7589) (5.9713, 4.9749) (4.7183, 10.8809) (8.4229, 13.0581) (4.3235, 11.3906) (4.7407, 8.0000) (2.0000, 1.7589) (2.0000, 1.5840) (3.3428, 2.8284) (2.0000, 1.3439) (2.0000, 2.6863) (22.2715, 33.7625) (73.0011, 52.6516) (2.3004, 2.5490) (2.0000, 2.1112) (12.2237, 11.0675) (6.1529, 6.4976) (17.3119, 20.7302) (6.1609, 6.9960) (7.8256, 6.7556) (14.1398, 9.5016) (8.3699, 8.8449) (24.0407, 19.9477) (9.9975, 7.8732) (4.1467, 7.8102) (2.0000, 1.0000) (4.9262, 9.5016) (7.8490, 6.4976) (38.3906, 32.2513) (109.7611, 70.1888) (3.0000, 1.0000) (4.1090, 2.6400) (2.4740, 1.7589) (4.1724, 9.7828) (3.2430, 3.0937) (2.4740, 1.7589) (2.0000, 1.7589) (2.9252, 1.9252) (16.1154, 15.6250) (6.0258, 9.5016) (241.5795, 160.8453) (2.0000, 1.1635) (3.1296, 6.9754) (18.0337, 20.2291) (125.2766, 160.8453) (19.0094, 22.6274) (3.2329, 2.8284) (11.2243, 11.1803) (48.7161, 38.3010) (2.0000, 6.3769) (10.0455, 9.5016) (74.0935, 50.5007) (27.6023, 24.0279) (2.5840, 1.7589) (12.8664, 12.3760) (3.7038, 3.3325) (2.8728, 2.8284) (2.1882, 1.9531) (5.6569, 3.0937) (2.1635, 1.7589) (3.0612, 2.8284) (4.5581, 3.8931) (4.5775, 5.1212) (3.5523, 3.5129) (5.1261, 3.0411) (3.5212, 5.0479) (3.2329, 3.2272) (5.2800, 5.0479) (8.0240, 8.2266) (3.5178, 2.0782) (3.3992, 1.9252) (24.0637, 20.1405) (41.6993, 18.2117) (24.1364, 18.3780) (4.2329, 1.9252) (2.5840, 1.9345) (3.2329, 2.2701) (18.1854, 14.6969) (7.8230, 11.1803) (5.1293, 2.4799) (5.6569, 2.8284) (2.0000, 1.7589) (8.0105, 6.4976) (5.6988, 3.8931) (8.6729, 8.9819) (2.0000, 1.0000) (2.4740, 2.3704) (9.4013, 6.0798) (2.0000, 2.2515) (4.3025, 3.6619) (2.2697, 1.7589) (2.4740, 1.7589) (7.1880, 11.2883) (12.3731, 13.4557) (2.7589, 3.3750) (10.4222, 17.0590) (11.4710, 12.8586) (3.5490, 2.5490) (4.1141, 7.2098) (12.3258, 6.0798) (6.9810, 4.9749) (2.4740, 1.8575) (38.0019, 54.5954) (4.3025, 3.8012) (2.0000, 1.0000) (9.9658, 6.4976) (27.9449, 20.2291) (331.8417, 695.5865) (99.4589, 216.8117) (18.4606, 227.4901) (7.4183, 5.0847) (2.0000, 1.0000) (5.2201, 6.4976) (7.0514, 6.0798) (5.1395, 2.8284) (11.4472, 3.9245) (2.7589, 3.1856) (2.2697, 1.7589) (2.0000, 1.7589) (107.4496, 184.4709) (4.2893, 15.2062) (2.4740, 1.9252) (60.5232, 148.6938) (395.8778, 201.0761) (2.0000, 2.3704) (15.5172, 13.9221) (3.2329, 1.7589) (2.7589, 1.7589) (2.0000, 5.3594) (31.5342, 70.7110) (4.2626, 3.7246) (21.4042, 17.8658) (16.9073, 7.1562) (4.5873, 6.0798) (146.8713, 198.0915) (3.4740, 1.4740) (11.5377, 36.0634) (5.1891, 3.8931) (2.7589, 7.7591) (12.0381, 6.2864) (48.1008, 158.6113) (2.9481, 6.0798) (3.0000, 6.1742) (2.6590, 1.4740) (110.4696, 118.7332) (78.8068, 60.8089) (3.7051, 2.0782) (5.8776, 4.9749) (4.8079, 7.8102) (20.5751, 27.0405) (5.7424, 15.9293) (2.0000, 1.7589) (8.8874, 16.2375) (2.0000, 4.6024) (4.5652, 3.2186) (479.2574, 332.1112) (279.9136, 249.6111) (3.1920, 13.8480) (2.7589, 10.2308) (76.2047, 176.6799) (3.2329, 2.8284) (370.9137, 415.5635) (3.0000, 1.4740) (12.0986, 9.4303) (2.0000, 1.7589) (296.5001, 262.5281) (8.0046, 7.1897) (4.9067, 2.8284) (2.4740, 4.4784) (4.3025, 3.8931) (33.2356, 21.0087) (3.8371, 2.8284) (19.9446, 13.7412) (28.0936, 16.6260) (4.5498, 3.8931) (6.0141, 3.7246) (4.5296, 3.1856) (3.0000, 2.5490) (2.9806, 2.0782) (9.6795, 11.1803) (3.4203, 2.0782) (2.0000, 1.0000) (6.4182, 6.9754) (28.8782, 64.2212) (6.0515, 5.1025) (20.3949, 11.1803) (7.6607, 9.5016) (2.7437, 2.0782) (3.5523, 2.8284) (2.0000, 1.6327) (2.8180, 2.8284) (9.3011, 9.1394) (2.0000, 1.0000) (2.3439, 1.4740) (5.6569, 3.8931) (9.0982, 5.2865) (4.1724, 4.2490) (53.1772, 29.2505) (2.3439, 1.4740) (3.2329, 1.7589) (473.7490, 419.9042) (41.5211, 16.2375) (3.4740, 1.4740) (13.8900, 11.1803) (5.5362, 6.6811) (8.7920, 12.3506) (11.5372, 9.9147) (99.6613, 73.1853) (20.4802, 39.9596) (2.5657, 1.4740) (19.2768, 26.7620) (2.0000, 4.0622) (4.6597, 3.1856) (312.7778, 304.2759) (17.8878, 107.2499) (6.9556, 4.9749) (54.7547, 52.0457) (2.0000, 5.0524) (21.1243, 19.4253) (48.0830, 46.8722) (53.9485, 34.5168) (2.0000, 1.4740) (2.4740, 1.7589) (40.8584, 19.4253) (2.0000, 1.3439) (47.4719, 34.5168) (2.9806, 9.7739) (2.9279, 1.7589) (2.0000, 1.2217) (2.9481, 1.4740) (2.0000, 1.0000) (2.0000, 1.0000) (40.0509, 40.0323) (7.2995, 4.1692) (106.1971, 239.9123) (2.6556, 1.7589) (2.0000, 1.5840) (2.0000, 1.2697) (7.1095, 10.0411) (12.2845, 17.5866) (2.7589, 6.6551) (15.9348, 19.2888) (18.6714, 15.4544) (2.0000, 1.3439) (2.0000, 1.7589) (47.0222, 36.0278) (6.8617, 9.5016) (2.0000, 1.3439) (2.0000, 2.8153) (2.4740, 6.4516) (5.9713, 6.7044) (2.4740, 1.7589) (17.2601, 21.9390) (41.1524, 32.3901) (2.6186, 6.0798) (2.0000, 1.4740) (5.1167, 14.2777) (5.4963, 6.0798) (3.1028, 3.2272) (3.2515, 2.5490) (5.8028, 3.7246) (87.3955, 76.3086) (64.8547, 33.9949) (14.6178, 34.2138) (9.9618, 6.0798) (7.6871, 13.0509) (2.7427, 6.7044) (609.3805, 554.2837) (44.9094, 34.6537) (4.9067, 2.8284) (4.1724, 7.8992) (5.0191, 5.3823) (92.1664, 92.6312) (2.0000, 29.9200) (2.9481, 3.1856) (4.2051, 2.8284) (4.0981, 5.4685) (3.6400, 11.3311) (5.8497, 6.4976) (3.2665, 2.0782) (4.1565, 2.6400) (2.0000, 1.7589) (68.7653, 36.1790) (2.0000, 1.9252) (7.0258, 3.8931) (20.7755, 16.2375) (2.0000, 1.4740) (2.0000, 5.3168) (9.4507, 19.4253) (2.7589, 2.3599) (2.0000, 1.0000) (4.1188, 3.7325) (2.1882, 15.8578) (7.0000, 4.6296) (33.9613, 77.1227) (3.1576, 2.2515) (2.8180, 2.2589) (15.1753, 11.1803) (4.2956, 2.6400) (72.1172, 109.9909) (5.8439, 3.8931) (3.7589, 6.1141) (2.8180, 4.7118) (2.0000, 1.0000) (3.9840, 2.8284) (3.5523, 2.8284) (4.1164, 5.1962) (4.4656, 4.0009) (3.5490, 7.5342) (6.0613, 2.8284) (3.2329, 2.0782) (3.5178, 2.8284) (2.0000, 1.4740) (3.5178, 1.7589) (2.9481, 1.7589) (2.4740, 2.0782) (2.0000, 11.7287) (2.6958, 1.7589) (2.0000, 1.9462) (2.0000, 1.4740) (2.4740, 4.6523) (3.2329, 2.0782) (2.0000, 1.3439) (3.7051, 2.8284) (2.4740, 3.3750) (3.7143, 4.2361) (4.4553, 4.0009) (4.7747, 2.8284) (5.5147, 4.7827) (9.8585, 9.5016) (3.5523, 2.8284) (2.5657, 2.8284) (2.0000, 1.4740) (2.7589, 2.5490) (2.0000, 1.1445) (6.7338, 5.7749) (14.0433, 9.5016) (2.0000, 1.6327) (43.1359, 39.3730) (6.5234, 2.8284) (2.4914, 2.8284) (5.1899, 3.5129) (2.4740, 1.9531) (2.9806, 2.6664) (5.1891, 9.1394) (2.0000, 1.4740) (36.9460, 28.3266) (2.0000, 1.4740) (70.3503, 124.3412) (73.3390, 63.6239) (9.9729, 6.6811) (22.6104, 24.3904) (2.4740, 1.7589) (21.5743, 18.3780) (3.4222, 3.5129) (14.7137, 12.2818) (22.6050, 41.8498) (7.2904, 8.0000) (2.9252, 4.2361) (25.4249, 24.3904) (3.5490, 3.3750) (13.0721, 19.0017) (307.2101, 202.6257) (114.6946, 82.7528) (4.0981, 4.1692) (4.9067, 2.8284) (5.1339, 6.2432) (374.8831, 372.4672) (4.3025, 4.1692) (5.7228, 7.8102) (35.4531, 38.9707) (707.5331, 571.2249) (3.3000, 6.0798) (2.0000, 3.4850) (695.5661, 607.0680) (336.5953, 243.4949) (13.6791, 13.6488) (3.3704, 14.6195) (2.4740, 1.7589) (2.0000, 1.3439) (2.8180, 1.4740) (10.4838, 8.0000) (5.1883, 7.0160) (186.2230, 237.2841) (6.9102, 7.5540) (4.9067, 4.8111) (3.5178, 9.1987) (13.9736, 11.1803) (514.1153, 286.7270) (3.0000, 4.0622) (564.2069, 266.9708) (2.4740, 2.8186) (2.0000, 1.9252) (19.9658, 39.7675) (14.3413, 48.5258) (243.4982, 249.3795) (54.1443, 83.8949) (2.9806, 2.8284) (6.4937, 9.5016) (75.6909, 91.9296) (107.4285, 68.8620) (65.3809, 74.7274) (69.2986, 161.8317) (2.0000, 1.0000) (4.7696, 6.4976) (2.3987, 1.4740) (97.9350, 190.0005) (5.4834, 3.8931) (87.8776, 81.9765) (2.6136, 1.7589) (77.2264, 60.8089) (386.1311, 281.7410) (38.6691, 28.9573) (54.4740, 41.3425) (422.2737, 306.9004) (175.6875, 288.1538) (26.2575, 41.4608) (107.7535, 137.4995) (21.1960, 45.7375) (279.4233, 252.4260) (2.4740, 2.8153) (70.3164, 127.7274) (7.4060, 5.1962) (21.7499, 42.3937) (4.6597, 3.8931) (21.6990, 28.9573) (2.7589, 3.3750) (132.2344, 121.2499) (117.6183, 69.5499) (67.6388, 85.5773) (3.8371, 11.9760) (4.1724, 2.8284) (3.6622, 5.3823) (6.4044, 24.7410) (3.1028, 2.8284) (11.6016, 15.2052) (2.0000, 1.0000) (2.3439, 3.7169) (4.2047, 3.9245) (32.8682, 20.2291) (5.9725, 7.6639) (4.0231, 4.7657) (2.2697, 3.1856) (4.5873, 46.9032) (4.4068, 26.8134) (15.4847, 22.3423) (7.0541, 11.1803) (14.1064, 11.1803) (25.5705, 37.6589) (3.2329, 3.0937) (4.4452, 4.9749) (3.0580, 3.6055) (10.3119, 35.8535) (16.1659, 19.7293) (15.7880, 12.8586) (2.4740, 3.3750) (2.6958, 2.8284) (2.0000, 1.7589) (2.4740, 5.9583) (6.3780, 7.1333) (20.3039, 38.7795) (2.3439, 1.9462) (9.0414, 21.8043) (2.2697, 1.7589) (2.0000, 1.7589) (7.1184, 5.5089) (3.5178, 6.4895) (2.1445, 1.9401) (5.0257, 8.6283) (2.4740, 7.6773) (4.0790, 2.8284) (2.0000, 1.7589) (9.4137, 12.8586) (8.6212, 17.9443) (8.9936, 11.1803) (2.5840, 2.3704) (3.4222, 2.0782) (214.5960, 216.7819) (38.4564, 92.1585) (3.0580, 6.3032) (2.0000, 2.4541) (2.0000, 1.2217) (55.1517, 122.4375) (2.0000, 1.7589) (2.3439, 1.9252) (3.3354, 6.9983) (2.0000, 1.0000) (2.0000, 1.0000) (4.5298, 6.6927) (73.6454, 101.8404) (4.2504, 12.8398) (59.0174, 60.8089) (2.1445, 1.3987) (23.8908, 35.0652) (2.7589, 12.8365) (3.5178, 4.5802) (5.3671, 51.9145) (2.0000, 1.7589) (5.0046, 3.3750) (4.3025, 4.9749) (2.0000, 1.3439) (16.6257, 16.2375) (2.8728, 2.8284) (3.8371, 2.5490) (6.5331, 3.8931) (2.0000, 3.7246) (4.0981, 5.7749) (2.4740, 1.7589) (7.3257, 6.1572) (6.8894, 5.4454) (2.2697, 1.4740) (2.0000, 1.4740)
    };
    \addlegendentry{Frangi$+$Sato ($\alpha = 1.81 \pm 0.03$, $R^2 = 0.761$)}
    
    \end{loglogaxis}
    \end{tikzpicture}
    \caption{Murray's law validation for the three vesselness
    weight configurations. Each marker represents a single
    bifurcation, plotting $r_{\mathrm{parent}}^{3}$ against
    $\sum r_{\mathrm{child},i}^{3}$. The black dashed line is the
    classical cube law ($\alpha = 3$). The gray dash-dotted central
    line and surrounding shaded band show the empirical  range $\alpha = 2.31 \pm 0.6$
    reported for human pulmonary vasculature, mapped to the $r^{3}$
    axes under the assumption of symmetric dichotomous bifurcations.
    The fitted $\alpha$ (point estimate as defined in Section~\ref{subsec:murray} $\pm$ bootstrap standard
    error, $n=1000$ resamples) and $R^{2}$ are given in the legend.}
    \label{fig:murray_sweep}
\end{figure}

The bounded nonlinear minimization of~\eqref{eq:murray_objective} over all bifurcation nodes of degree~$\geq 3$ (Fig.~\ref{fig:murray_sweep}) gave $\alpha~= 2.81~\pm~0.03$ with $R^2~= 0.753$ for Frangi, $\alpha~= 1.76~\pm~0.03$ with $R^2~= 0.760$ for Sato, and $\alpha~= 1.81~\pm~0.03$ with $R^2~=~0.761$ for the fused configuration. Errors are nonparametric bootstrap standard errors from $n~=~1000$ resamples.

\subsubsection{Horton Ratios}
Pooled across consecutive Strahler-order pairs from both lungs, the bifurcation ratio was $R_B~=~3.68~\pm~2.16$ (Frangi), $3.24~\pm~1.05$ (Sato), and $3.41~\pm~1.28$ (fusion). The length ratio was $R_L~= 2.17~\pm~1.59$, $1.61~\pm~1.33$, and $1.64~\pm~0.84$, and the diameter ratio was $R_D~= 1.10~\pm~0.14$, $1.54~\pm~0.32$, and $1.43~\pm~0.15$, respectively.

\subsubsection{Tortuosity}
Mean edge tortuosity was $\tau~= 1.13~\pm~0.16$ for Frangi, $1.16~\pm~0.21$ for Sato, and $1.16~\pm~0.20$ for the fused configuration, computed across all edges of both lungs.

Table~\ref{tab:morphometry} summarizes the morphometric parameters of the reconstructed vascular graph for each vesselness filter configuration.

\begin{table}[!t]
\centering
\caption{Morphometric comparison of the reconstructed vascular
graph across three vesselness filter configurations. 
Strahler order is aggregated across left
and right lungs, the remaining metrics are reported as
mean\,$\pm$\,SD pooled over both lungs: across all bifurcations
for $\delta$, across all edges for $\tau$ and across all
Strahler order pairs for the Horton ratios $R_B$, $R_L$, $R_D$.
The Murray exponent $\alpha$ is given as the point estimate as defined in Section~\ref{subsec:murray}
$\pm$ nonparametric bootstrap standard error. }
\label{tab:morphometry}
\renewcommand{\arraystretch}{1.15}
\begin{tabular}{@{} l c c c @{}}
\hline\hline
\textbf{Metric} & \textbf{Frangi} & \textbf{Sato} & \textbf{Frangi+Sato} \\
\hline
\multicolumn{4}{@{}l}{\rule{0pt}{2.4ex}\textbf{Graph topology}} \\
\quad Nodes & 8433 & 10113 & 12929 \\
\quad Bifurcations & 3351 & 3534 & 4520 \\
\noalign{\vskip 3pt}
\multicolumn{4}{@{}l}{\rule{0pt}{2.4ex}\textbf{Strahler ordering}} \\
\quad Max.\ order & 7 & 6 & 7 \\
\quad Asymmetry $\delta$ & 0.90 $\pm$ 0.10 & 0.87 $\pm$ 0.12 & 0.88 $\pm$ 0.12 \\
\noalign{\vskip 3pt}
\multicolumn{4}{@{}l}{\rule{0pt}{2.4ex}\textbf{Fractal dimension}} \\
\quad $D_f^{\mathrm{skel}}$ & 1.83 $\pm$ 0.11 & 1.84 $\pm$ 0.10 & 1.91 $\pm$ 0.10 \\
\quad $D_f^{\mathrm{vol}}$ & 2.12 $\pm$ 0.05 & 2.21 $\pm$ 0.04 & 2.26 $\pm$ 0.05 \\
\noalign{\vskip 3pt}
\multicolumn{4}{@{}l}{\rule{0pt}{2.4ex}\textbf{Murray's law}} \\
\quad $\alpha$ & 2.81 $\pm$ 0.03 & 1.76 $\pm$ 0.03 & 1.81 $\pm$ 0.03 \\
\noalign{\vskip 3pt}
\multicolumn{4}{@{}l}{\rule{0pt}{2.4ex}\textbf{Horton ratios}} \\
\quad $R_B$ (bifurcation) & 3.68 $\pm$ 2.16 & 3.24 $\pm$ 1.05 & 3.41 $\pm$ 1.28 \\
\quad $R_L$ (length) & 2.17 $\pm$ 1.59 & 1.61 $\pm$ 1.33 & 1.64 $\pm$ 0.84 \\
\quad $R_D$ (diameter) & 1.10 $\pm$ 0.14 & 1.54 $\pm$ 0.32 & 1.43 $\pm$ 0.15 \\
\noalign{\vskip 3pt}
\multicolumn{4}{@{}l}{\rule{0pt}{2.4ex}\textbf{Tortuosity}} \\
\quad $\tau$ & 1.13 $\pm$ 0.16 & 1.16 $\pm$ 0.21 & 1.16 $\pm$ 0.20 \\
\noalign{\vskip 3pt}
\hline\hline
\end{tabular}
\end{table}

\section{Discussion}
\label{sec:discussion}

\subsection{Interpretation of Morphometric Results}
\label{ssec:interpretation}

\subsubsection{Graph Topology}
The fused tree contains 12\,929 nodes and 4\,520 bifurcations - 53\% more nodes than Frangi alone (8\,433) and 28\% more than Sato (10\,113). Frangi loses large-vessel continuity, Sato misses distal branches, and fusion recovers both.
  
 \subsubsection{Strahler Ordering and Asymmetry Ratio}
  Frangi and fusion reach a maximum Strahler order of~7, while Sato alone
  reaches order~6, all well below the 11--15 orders reported from full
  resin casts~\cite{huang1996,burrowes2005}. The asymmetry ratio ranges from $0.87 \pm 0.12$ (Sato) to $0.90 \pm 0.10$ (Frangi), above the reference value $\delta \approx 0.67$ reported for
  pulmonary arterial dichotomies~\cite{koike1986}. Both differences come from distal branches ($d < 0.5$\,mm, $\delta \approx 0.1$\text{--}$0.3$), which
  account for 50--80\% of bifurcations in the intact tree. These branches fall below the CT resolution and are not resolved, which simultaneously truncates the deep Strahler orders and biases the observable $\delta$ toward the near-symmetric dichotomous subset.
  
 \subsubsection{Volumetric Fractal Dimension} $D_f^{\mathrm{vol}}~= 2.26~\pm~0.05$ for fusion falls within the 2.21--2.44 range reported for pulmonary vasculature at clinical CT resolution~\cite{helmberger2014}. Sato reaches the lower bound of this range, whereas Frangi alone underestimates space-filling complexity due to a sparser vascular tree.
  
\subsubsection{Murray's Law} 
All three configurations fall within the empirical range $\alpha~= 2.31~\pm~0.60$ reported for human pulmonary arteries~\cite{revisiting_murray2025}, supporting the broad plausibility of the reconstructed trees. Frangi ($\alpha~= 2.81~\pm~0.03$), which captures distal vasculature, is closest to the classical optimum $\alpha~= 3$~\cite{murray1926}. Sato ($\alpha = 1.76 \pm 0.03$) and fusion ($\alpha~= 1.81~\pm~0.03$), whose trees are richer in proximal bifurcations, yield lower exponents. This pattern aligns with the expectation that Murray's energy-minimization assumptions hold more closely in the distal vasculature. Nevertheless, a single, low-resolution scan is insufficient to confirm this interpretation.

\subsubsection{Horton Ratios}
The bifurcation ratio $R_B$ ranges from $3.24~\pm~1.05$ (Sato) to $3.68~\pm~2.16$ (Frangi), roughly aligning with the anatomical range of 3.03--3.36~\cite{burrowes2005}. The exceptionally large standard deviations stem from two practical artifacts: CT resolution truncates the distal networks, where Horton self-similarity is most stable, and single-digit vessel counts at the highest proximal Strahler orders cause large swings in the ratio. The length ratio $R_L$ follows a similar pattern ($1.61~\pm~1.33$ for Sato to $2.17~\pm~1.59$ for Frangi), exceeding the 1.49--1.52 reference~\cite{burrowes2005}, with Frangi's value inflated by numerous short peripheral segments. Conversely, the diameter ratio $R_D$ is more stable ($\sigma \leq 0.32$), indicating smooth caliber tapering. Sato ($1.54~\pm~0.32$) approaches the 1.56--1.60 reference~\cite{burrowes2005}, while fusion ($1.43~\pm~0.15$) and Frangi ($1.10~\pm~0.14$) show less diameter differentiation across orders.

\subsubsection{Tortuosity}
Mean tortuosity $\tau \approx 1.13\text{--}1.16$ with $\sigma_\tau~\leq~0.21$ is essentially invariant to filter choice, which confirms that the centerline geometry depends on the TEASAR skeletonization rather than the stage of vesselness enhancement.

\section{Limitations}
 \label{sec:limitations}

This study has three primary limitations. First, all results are derived from a single-patient dataset. Therefore, the generalizability of the reported morphometric values, graph reconstruction, and optimal parameter selection remains to be validated in a multi-patient cohort. Second, the absence of a ground-truth reference prevented a comprehensive assessment of reconstruction quality. Finally, the SPECT/CT scanner's limited spatial resolution renders a portion of the pulmonary vasculature undetectable, compromising reconstruction completeness and introducing bias into the derived morphometric indices.

\section{Conclusion}
\label{sec:conclusion}

  This work presented a preliminary pipeline for pulmonary vascular tree reconstruction from a clinical non-contrast chest CT scan done on a SPECT/CT scanner, which offers substantially lower spatial resolution and higher noise levels than a dedicated diagnostic CT. Despite these input conditions, the pipeline produced morphometrically plausible vascular trees by combining vesselness enhancement, TEASAR skeletonization, and graph-based morphometric analysis, the results of which provide a meaningful indirect validation of the reconstructed vascular tree.
  
  The weighted fusion configuration
  ($w_{\mathrm{Frangi}}=1.0$, $w_{\mathrm{Sato}}=~1.2$) produced the richest vascular graph, recovering 53\% more vascular nodes than Frangi alone and 28\% more than Sato alone. The resulting graph yielded a
  volumetric fractal dimension $D_f^{\mathrm{vol}} = 2.26 \pm 0.05$
  within the literature range $2.21\text{--}2.44$, a Murray exponent
  $\alpha = 1.81 \pm 0.03$ within the empirical pulmonary range
  $2.31 \pm 0.60$, and Horton ratios
  ($R_B = 3.41 \pm 1.28$, $R_L = 1.64 \pm 0.84$,
  $R_D = 1.43 \pm 0.15$) roughly aligned with the established anatomical norms.

\section*{Code Availability}
The source code used to implement the deterministic pulmonary vascular tree reconstruction pipeline is available at: \url{https://github.com/rroszczyk/PulmoVesselGraph}. The repository includes scripts for CT preprocessing, multiscale vesselness filtering, vascular skeletonization, graph construction, and morphometric analysis.

\printbibliography

\end{document}